\documentclass[ usenames,dvipsnames]{article} 
\usepackage{iclr2022_conference,times}

\usepackage{hyperref}
\usepackage{url}
\usepackage{amssymb}
\usepackage{amsmath}
\usepackage{booktabs}
\usepackage{enumitem}
\usepackage{graphicx}
\usepackage{svg}
\usepackage{float}
\usepackage{wrapfig}
\usepackage{subcaption}
\usepackage{epstopdf}
\usepackage{appendix}
\usepackage{epsf}
\usepackage{svg}

\usepackage{fancyheadings}
\usepackage{graphics}
\usepackage[normalem]{ulem}
\usepackage{enumitem}
\usepackage{booktabs}
\usepackage{nicefrac}
\usepackage{subcaption}
\usepackage{psfrag}
\usepackage{comment}
\usepackage{color}
\usepackage{wrapfig}
\usepackage{pgfplots}
\usepackage{pgfplotstable}
\usepackage{multirow}
\usepackage{mathtools}
\usepackage{amsfonts}
\usepackage{amsthm}
\usepackage{amsmath}
\usepackage{amssymb}
\usepackage{scalerel}
\usepackage{nccmath}
\usepackage{tikz}
\usepackage{algorithmic}
\usepackage[ruled,vlined]{algorithm2e}

\usepackage{hyperref}
\hypersetup{colorlinks, linkcolor = red, urlcolor  = YellowOrange, citecolor = blue, anchorcolor = YellowOrange}

\newcommand{\trafficm}{Traffic-Mer}
\newcommand{\trafficb}{Traffic-Bot}

\definecolor{bluetext}{HTML}{1F77B4}
\definecolor{greentext}{HTML}{2CA02C}
\definecolor{purpletext}{HTML}{663399}
\definecolor{yellowtext}{HTML}{FFC62F}

\title{The Effects of Reward Misspecification: \\Mapping and Mitigating Misaligned Models}

\author{Alexander Pan\\Caltech \And Kush Bhatia\\UC Berkeley \And Jacob Steinhardt\\UC Berkeley
}
\iclrfinalcopy 
\begin{document}

\maketitle
\begin{abstract}
    Reward hacking---where RL agents exploit gaps in misspecified reward functions---has been widely observed, but not yet systematically studied. To understand how reward hacking arises, we construct four RL environments with misspecified rewards. 
    We investigate reward hacking as a function of agent capabilities: model capacity, action space resolution, observation space noise, and training time. More capable agents often exploit reward misspecifications, achieving higher proxy reward and lower true reward than less capable agents. Moreover, we find instances of \emph{phase transitions}: capability thresholds at which the agent's behavior qualitatively shifts, leading to a sharp decrease in the true reward. 
    Such phase transitions pose challenges to monitoring the safety of ML systems. To 
    address this, we propose an anomaly detection task for aberrant policies and offer several baseline detectors.
\end{abstract}
\section{Introduction}\label{sec:intro}

As reinforcement learning agents are trained with better algorithms, more data, and larger policy models, they are at increased risk of overfitting their objectives \citep{russell2019human}. \emph{Reward hacking}, or the gaming of misspecified reward functions by RL agents, has appeared in a variety of contexts, such as game playing~\citep{ibarz2018humandemo}, 
text summarization~\citep{paulus2018drl_summarization}, and autonomous driving~\citep{knox2021misdesign}. 
These examples show that better algorithms and models are not enough; for human-centered applications such as healthcare~\citep{yu2019rlhealth}, economics~\citep{trott2021aieconomist} and  robotics~\citep{kober2013robotics}, RL algorithms must be safe and aligned with human objectives~\citep{bommansani2021foundational,hubinger2019risks}. 

Reward misspecifications occur because real-world tasks have numerous, often conflicting desiderata. In practice, reward designers resort to optimizing a proxy reward that is either more readily measured or more easily optimized than the true reward. For example, 
consider a recommender system optimizing for users' subjective well-being (SWB). Because SWB is difficult to measure, engineers rely on more tangible metrics such as click-through rates or watch-time. Optimizing for misspecified proxies led YouTube to overemphasize watch-time and harm user satisfaction~\citep{stray2020recommendersystem}, as well as to recommended extreme political content to users~\citep{times2019brazil}. 

Addressing reward hacking is a first step towards developing human-aligned RL agents and one goal of ML safety~\citep{Hendrycks2021UnsolvedPI}. However, there has been little systematic work investigating when or how it tends to occur, or how to detect it before it runs awry. To remedy this, we study the problem of reward hacking across four diverse environments: traffic control~\citep{wu2017flow}, COVID response~\citep{kompella2020pandemic}, blood glucose monitoring~\citep{fox2020bgp}, and the Atari game Riverraid~\citep{brockman2016gym}. Within these environments, we construct nine misspecified proxy reward functions (Section~\ref{sec:setup}).  

\begin{figure}[tbp]
\centering
\includegraphics[scale=0.21]{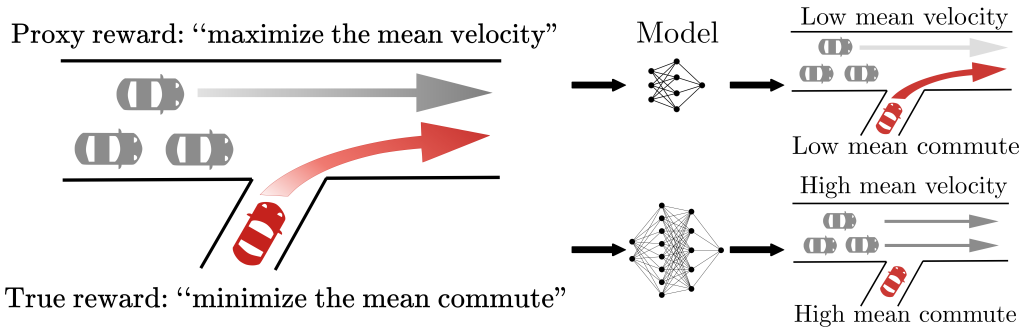}%




\caption{An example of reward hacking when cars merge onto a highway. A human-driver model controls the grey cars and an RL policy controls the red car. The RL agent observes positions and velocities of nearby cars (including itself) and adjusts its acceleration to maximize the proxy reward. At first glance, both the proxy reward and true reward appear to incentivize fast traffic flow. However, smaller policy models allow the red car to merge, whereas larger policy models exploit the misspecification by stopping the red car. When the red car stops merging, the mean velocity increases (merging slows down the more numerous grey cars). However, the mean commute time also increases (the red car is stuck). This exemplifies a \emph{phase transition}: the qualitative behavior of the agent shifts as the model size increases.}
\label{fig:main}
\end{figure}

Using our environments, we study how increasing optimization power affects reward hacking, by training RL agents with varying resources such as model size, training time, action space resolution, and observation space noise (Section~\ref{sec:measurement}). We find that more powerful agents often 
attain higher proxy reward but lower true reward, as illustrated in Figure~\ref{fig:main}. Since the trend in ML is to increase resources exponentially each year \citep{ai100report2021}, this suggests that reward hacking will become more pronounced in the future in the absence of countermeasures.




More worryingly, we observe several instances of \emph{phase transitions}. In a phase transition, the more capable model pursues a qualitatively different policy that sharply decreases the true reward. Figure~\ref{fig:main} illustrates one example: An RL agent regulating traffic learns to stop any cars from merging onto the highway in order to maintain a high average velocity of the cars on the straightaway. 

Since there is little prior warning of phase transitions, they pose a challenge to monitoring the safety of ML systems.
Spurred by this challenge, we propose an anomaly detection task~\citep{Hendrycks2017ABF, 
Tack2020CSIND}: Can we detect when the true reward starts to drop, while maintaining a low false positive rate in benign cases? We instantiate our proposed task, \textsc{Polynomaly}, for the traffic and COVID environments (Section~\ref{sec:detection}). Given a trusted policy with moderate performance, one must detect whether a given policy is aberrant. 
We provide several baseline anomaly detectors for this task and release our data at \url{https://github.com/aypan17/reward-misspecification}.

\section{Related Work}\label{sec:related}
Previous works have focused on classifying different types of reward hacking and sometimes mitigating its effects. One popular setting is an agent on a grid-world with an erroneous sensor. \citet{hadfield2017ird} show and mitigate the reward hacking that arises due to an incorrect sensor reading at test time in a 10x10 navigation grid world. \citet{leike2017gridworld} show examples of reward hacking in a 3x3 boat race and a 5x7 tomato watering grid world. \citet{everitt2017corruptedreward} theoretically study and mitigate reward hacking caused by a faulty sensor.  

Game-playing agents have also been found to hack their reward. \citet{baker2020emergent} exhibit reward hacking in a hide-and-seek environment comprising 3-6 agents, 3-9 movable boxes and a few ramps: without a penalty for leaving the play area, the hiding agents learn to endlessly run from the seeking agents. \citet{toromanoff2019drl} briefly mention reward hacking in several Atari games (Elevator Action, Kangaroo, Bank Heist) where the agent loops in a sub-optimal trajectory that provides a repeated small reward. 

Agents optimizing a learned reward can also demonstrate reward hacking. \citet{ibarz2018humandemo} show an agent hacking a learned reward in Atari (Hero, Montezuma's Revenge, and Private Eye), where optimizing a frozen reward predictor eventually achieves high predicted score and low actual score. \citet{christiano2017preflearning} show an example of reward hacking in the Pong game where the agent learns to hit the ball back and forth instead of winning the point. \citet{stiennon2020learning} show that a policy which over-optimizes the learnt reward model for text summarization produces lower quality summarizations when judged by humans.

\section{Experimental Setup: Environments and Reward Functions} \label{sec:setup}
In this section, we describe our four environments (Section~\ref{sec:environments}) and taxonomize our nine corresponding misspecified reward functions (Section~\ref{sec:misspec}).


\subsection{Environments}\label{sec:environments}
We chose a diverse set of environments and prioritized complexity of action space, observation space, and dynamics model. Our aim was to reflect real-world constraints in our environments, selecting ones with several desiderata that must be simultaneously balanced.  Table~\ref{tab:misspecification_summary} provides a summary.



\paragraph{Traffic Control.}
The traffic environment is an autonomous vehicle (AV) simulation that models vehicles driving on different highway networks. The vehicles are either controlled by a RL algorithm or pre-programmed via a human behavioral model. Our misspecifications are listed in Table~\ref{tab:misspecification_summary}.

We use the Flow traffic simulator, implemented by~\citet{wu2017flow} and~\citet{vinitsky2018flowbench}, which extends the popular SUMO traffic simulator~\citep{sumo2018}. The simulator uses cars that drive like humans, following the Intelligent Driver Model (IDM)~\citep{treiber2000idm}, a widely-accepted approximation of human driving behavior. Simulated drivers attempt to travel as fast as possible while tending to decelerate whenever they are too close to the car immediately in front. 


The RL policy has access to observations only from the AVs it controls. For each AV, the observation space consists of the car's position, its velocity, and the position and velocity of the cars immediately in front of and behind it. The continuous control action 
is the acceleration applied to each AV. 
Figure~\ref{fig:traffic_qualitative} depicts the \trafficm~network, where cars from an on-ramp attempt to merge onto the straightaway. 
We also use the \trafficb~network, where cars (1-4 RL, 10-20 human) drive through a highway bottleneck where lanes decrease from four to two to one.

\paragraph{COVID Response.}
The COVID environment, developed by~\citet{kompella2020pandemic}, simulates a population using the SEIR model of individual infection dynamics. The RL policymaker adjusts the severity of social distancing regulations while balancing economic health (better with lower regulations) and public health (better with higher regulations), similar in spirit to~\citet{trott2021aieconomist}. The population attributes (proportion of adults, number of hospitals) and infection dynamics (random testing rate, infection rate) are based on data from Austin, Texas.

Every day, the environment simulates the infection dynamics and reports testing results to the agent, but not the true infection numbers. The policy chooses one of three discrete actions: \textsc{increase}, \textsc{decrease}, or \textsc{maintain} the current regulation stage, which directly affects the behavior of the population and indirectly affects the infection dynamics. There are five stages in total. 



\paragraph{Atari Riverraid.}
The Atari Riverraid environment is run on OpenAI Gym~\citep{brockman2016gym}. 
The agent operates a plane which flies over a river and is rewarded by destroying enemies. 
The agent observes the raw pixel input of the environment. The agent can take one of eighteen discrete actions, corresponding to either movement or shooting within the environment. 

\paragraph{Glucose Monitoring.}
The glucose environment, implemented in~\citet{fox2020bgp}, is a continuous control problem. It extends a FDA-approved simulator~\citep{man2014diabetes} for blood glucose levels of a patient with Type 1 diabetes. The patient partakes in meals and wears a continuous glucose monitor (CGM), which gives noisy observations of the patient's glucose levels. The RL agent administers insulin to maintain a healthy glucose level.

Every five minutes, the agent observes the patient's glucose levels and decides how much insulin to administer. The observation space is the previous four hours of glucose levels and insulin dosages. 


\subsection{Misspecifications}
\label{sec:misspec}

Using the above environments, we constructed nine instances of misspecified proxy rewards. To help interpret these proxies, 
we taxonomize them as instances of misweighting, incorrect ontology, or incorrect scope. We elaborate further on this taxonimization using the traffic example from Figure~\ref{fig:main}. 
%
\begin{itemize}[leftmargin=*]
    \item \textbf{Misweighting.} Suppose that the true reward is a linear combination of commute time and acceleration (for reducing carbon emissions). Downweighting the acceleration term thus underpenalizes carbon emissions. In general, misweighting occurs when the proxy and true reward capture the same desiderata, but differ on their relative importance.
    \item \textbf{Ontological.} Congestion could be operationalized as either high average commute time or low average vehicle velocity. In general, ontological misspecification occurs when the proxy and true reward use different desiderata to capture the same concept. 
    \item \textbf{Scope.} If monitoring velocity over all roads is too costly, a city might instead monitor them only over highways, thus pushing congestion to local streets. In general, scope misspecification occurs when the proxy measures desiderata over a restricted domain (e.g.~time, space).  
\end{itemize}
We include a summary of all nine tasks in  Table~\ref{tab:misspecification_summary} and provide full details in Appendix~\ref{app:measurement}. Table~\ref{tab:misspecification_summary} also indicates whether each proxy leads to misalignment (i.e.~to a policy with low true reward) and whether it leads to a phase transition (a sudden qualitative shift as model capacity increases). We investigate both of these in Section~\ref{sec:measurement}. 

{\setlength{\tabcolsep}{5pt}
\begin{table}[tbp]
\centering
\scalebox{0.9}{
\begin{tabular}{cccccc}
\toprule
\textbf{Env.} &Type& \makebox[5em]{Objective}&Proxy & Misalign? & Transition?\\
\midrule
\multirow{4}{*}{Traffic} & \multirow{1}{*}{Mis.} & \multirow{4}{*}{\shortstack[l]{minimize commute \\ and accelerations}}  & \multirow{1}{*}{underpenalize acceleration} &\multirow{1}{*}{No} &\multirow{1}{*}{No}\\
 & \multirow{1}{*}{Mis.}&  & \multirow{1}{*}{underpenalize lane changes} &\multirow{1}{*}{Yes} & \multirow{1}{*}{Yes}\\
 & \multirow{1}{*}{Ont.} &  & \multirow{1}{*}{velocity replaces commute} &\multirow{1}{*}{Yes} & \multirow{1}{*}{Yes}\\
 & \multirow{1}{*}{Scope} &  & \multirow{1}{*}{monitor velocity  near merge} & \multirow{1}{*}{Yes} & \multirow{1}{*}{Yes}\\
 \midrule 
\multirow{3}{*}{COVID} & \multirow{2}{*}{Mis.} & \multirow{3}{*}{\shortstack[l]{balance economic, \\ health, political cost}} & \multirow{2}{*}{underpenalize health cost}& \multirow{2}{*}{No} & \multirow{2}{*}{No}\\
 & \multirow{2}{*}{Ont.} &  & \multirow{2}{*}{ignore political cost} & \multirow{2}{*}{Yes} & \multirow{2}{*}{Yes}\\
 & & & & &\\
 \midrule 
\multirow{3}{*}{Atari} & \multirow{2}{*}{Mis.} & \multirow{3}{*}{\shortstack[l]{score points under\\smooth movement }}   &  \multirow{2}{*}{downweight movement}&\multirow{2}{*}{No} & \multirow{2}{*}{No}\\
 & \multirow{2}{*}{Ont.} &  & \multirow{2}{*}{include shooting penalty}& \multirow{2}{*}{No} & \multirow{2}{*}{No}\\
 &&&&&\\
 \midrule 
Glucose & Ont. & minimize health risk & risk in place of cost & Yes & No\\

\bottomrule
\end{tabular}}
\caption{Reward misspecifications across our four environments. `Misalign' indicates whether the true reward drops and `Transition' indicates whether this corresponds to a phase transition (sharp qualitative change). We observe 5 instances of misalignment and 4 instances of phase transitions. `Mis.' is a misweighting and 'Ont.' is an ontological misspecification.}
\label{tab:misspecification_summary}
\end{table}}

\paragraph{Evaluation protocol.} For each environment and proxy-true reward pair, we train an agent using the proxy reward and evaluate performance according to the true reward. We use PPO~\citep{schulman2017ppo} to optimize policies for the traffic and COVID environments, SAC~\citep{haarnoja2018sac} to optimize the policies for the glucose environment, and  torchbeast~\citep{torchbeast2019}, a PyTorch implementation of IMPALA~\citep{espeholt2018impala}, to optimize the policies for the Atari environment. When available, we adopt the hyperparameters (except the learning rate and network size) given by the original codebase.



\section{How Agent Optimization Power Drives Misalignment}\label{sec:measurement}

\begin{figure}[bpt]
\centering
\subfloat[Traffic - Ontological] {\includegraphics[scale=0.41]{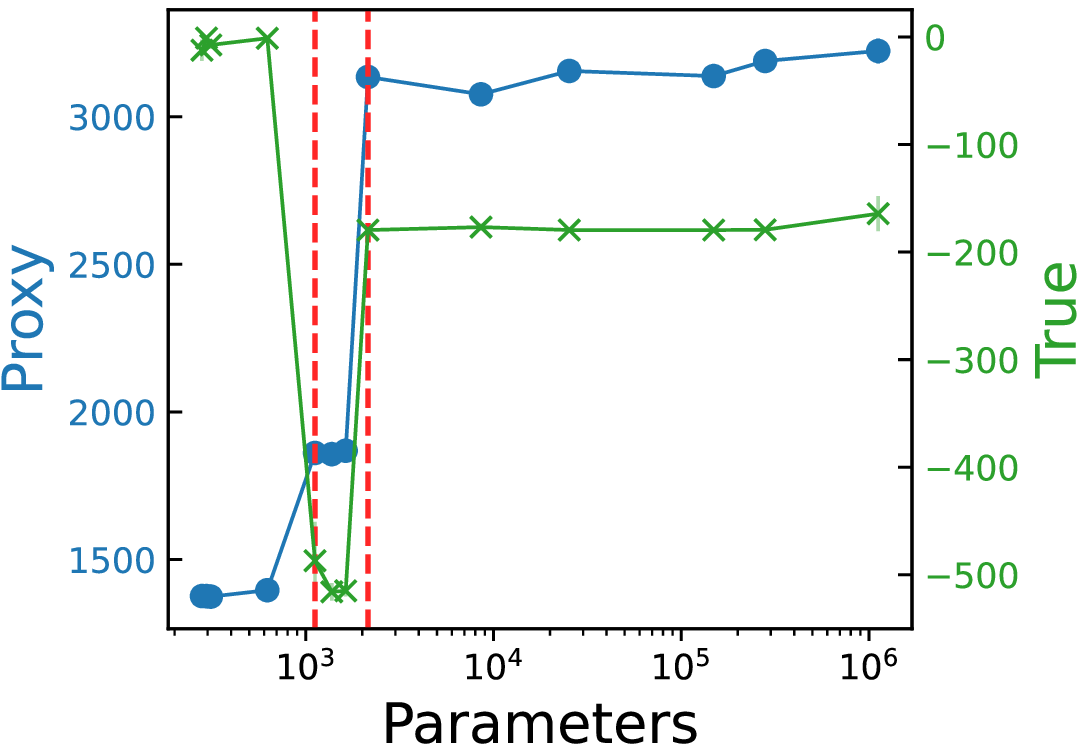}\label{fig:t_ont}}%
\hfill
\subfloat[COVID - Ontological] {\includegraphics[scale=0.41]{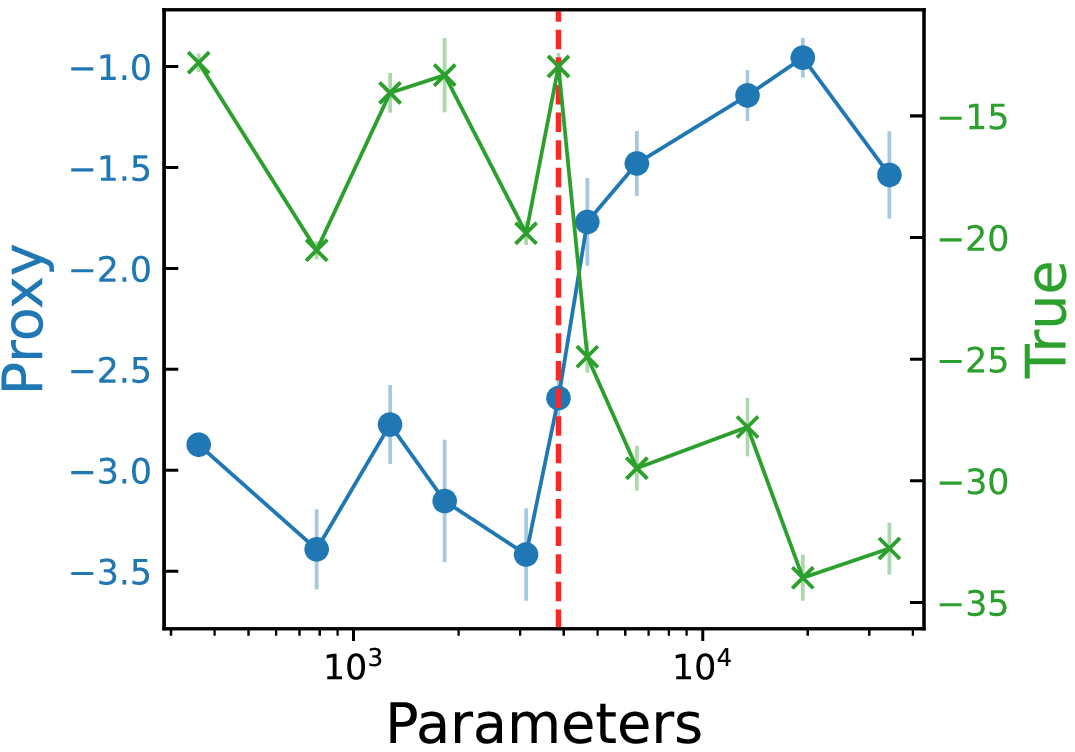}\label{fig:p_ont}}%
\hfill
\subfloat[Glucose - Ontological] {\includegraphics[scale=0.41]{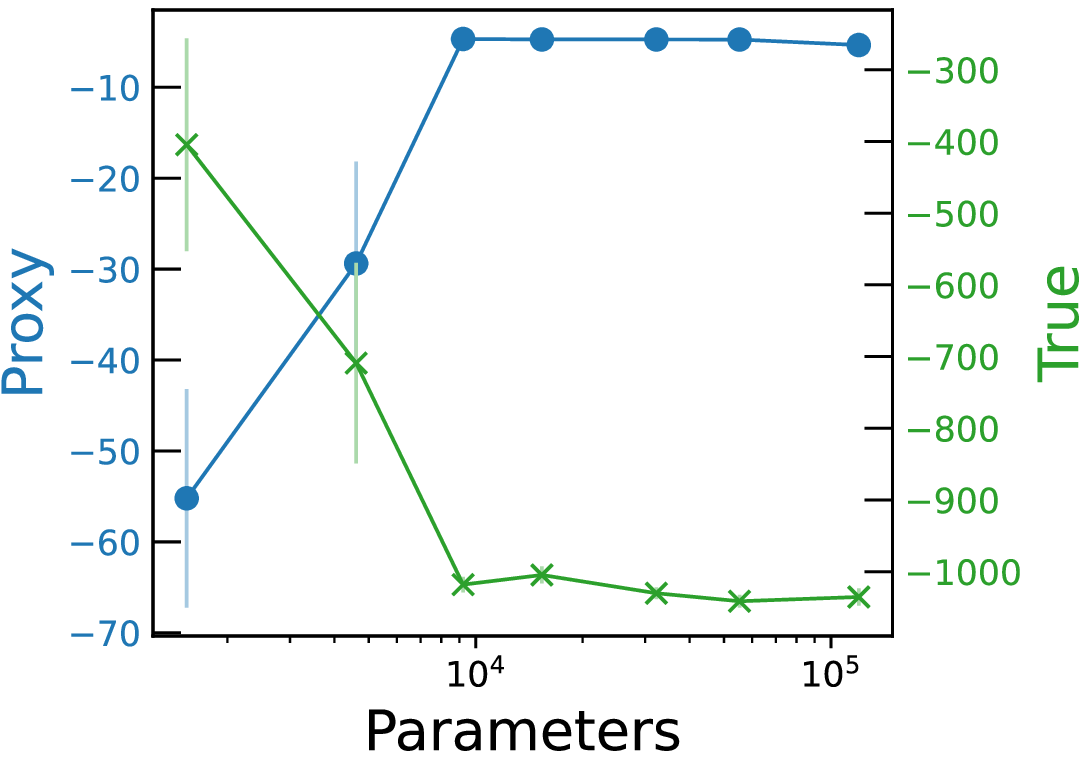}\label{fig:g_ont}}%

\caption{Increasing the RL policy's model size decreases true reward on three selected environments. The red line indicates a phase transition.}
\label{fig:modelsize_scatter}
\end{figure}

To better understand reward hacking, we study how it emerges as agent optimization power increases. We define optimization power as the effective search space of policies the agent has access to, as implicitly determined by model size, training steps, action space, and observation space.


In Section~\ref{sec:quantitative}, we consider the quantitative effect of optimization power for all nine environment-misspecification pairs; we primarily do this by varying model size, but also use training steps, action space, and observation space as robustness checks. Overall, more capable agents tend to overfit the proxy reward and achieve a lower true reward. We also find evidence of phase transitions on four of the environment-misspecification pairs. For these phase transitions, there is a critical threshold at which the proxy reward rapidly increases and the true reward rapidly drops.


 
 In Section~\ref{sec:qualitative}, we further investigate these phase transitions by qualitatively studying the resulting policies. At the transition, we find that the quantitative drop in true reward corresponds to a qualitative shift in policy behavior. Extrapolating visible trends is therefore insufficient to catch all instances of reward hacking, increasing the urgency of research in this area.
 
 In Section~\ref{sec:correlation}, we assess the faithfulness of our proxies, showing that reward hacking occurs even though the true and proxy rewards are strongly positively correlated in most cases.

\subsection{Quantitative Effects vs. Agent Capabilities}\label{sec:quantitative}

As a stand-in for increasing agent optimization power, we first 
vary the model capacity for a fixed environment and proxy reward. Specifically, we vary the width and depth of the actor and critic networks, changing the parameter count by two to four orders of magnitude depending on the environment. For a given policy, the actor and critic are always the same size.

 \begin{figure}[!bp]
    \centering
  \subfloat[Atari - Misweighting\label{fig:training_steps}]{\includegraphics[scale=0.27]{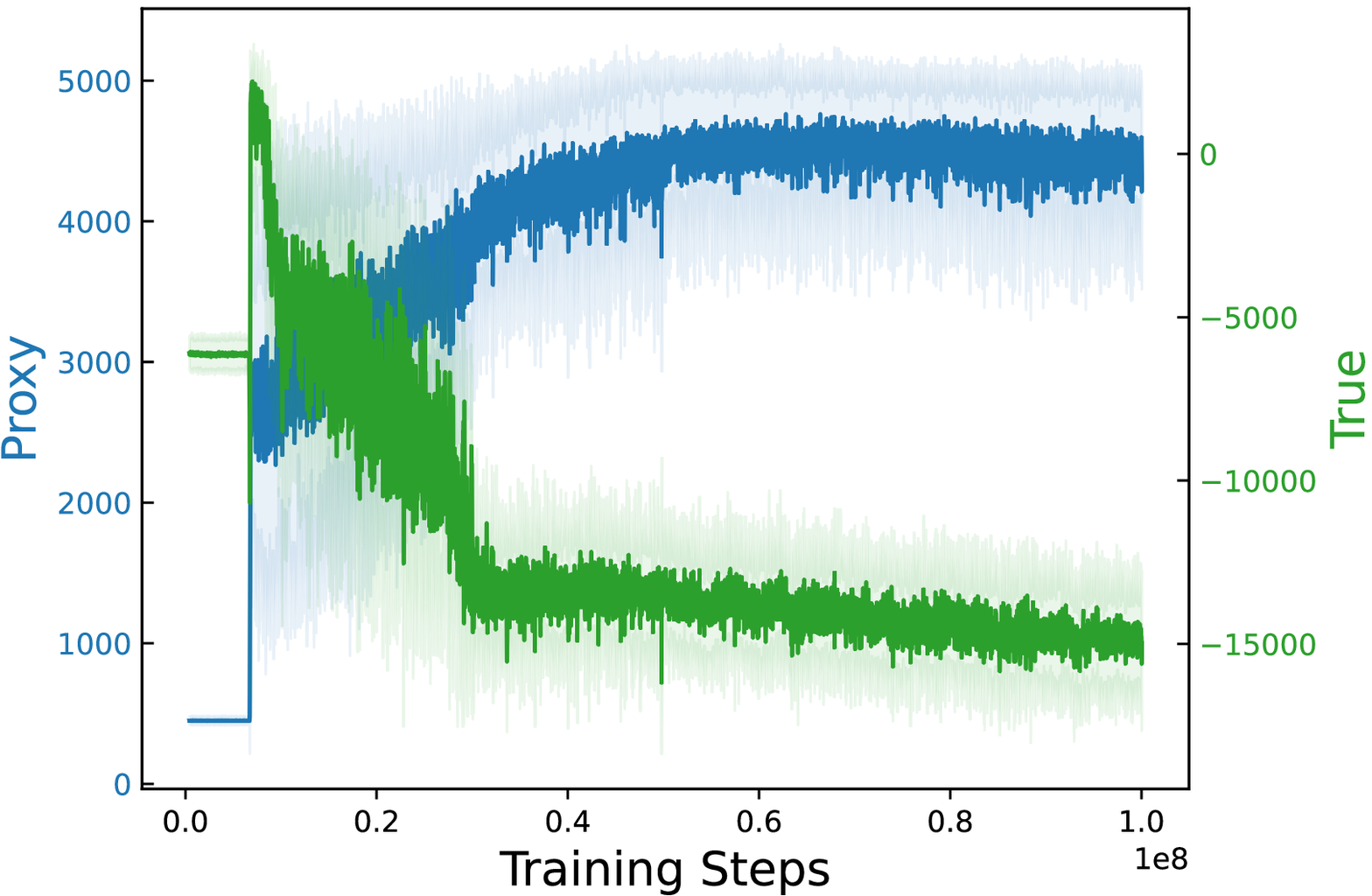}}%
  \hfill
  \subfloat[Traffic - Ontological\label{fig:action_noise}] {\includegraphics[scale=0.4]{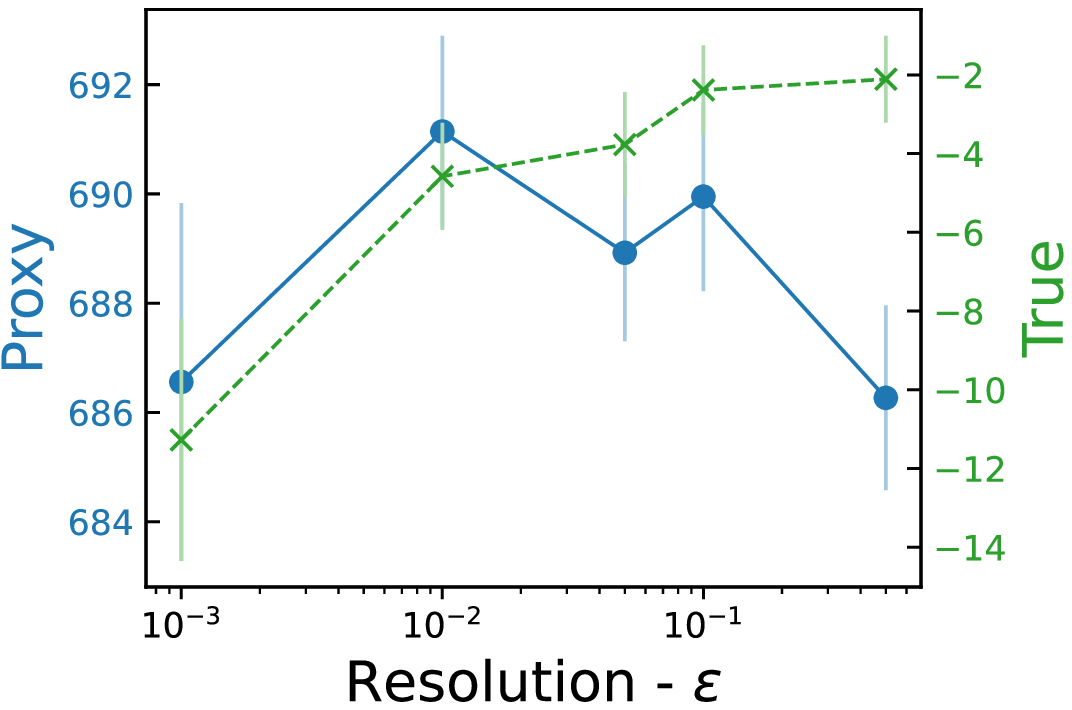}}%
  \hfill \subfloat[COVID - Ontological\label{fig:obs_noise}]{\includegraphics[scale=0.38]{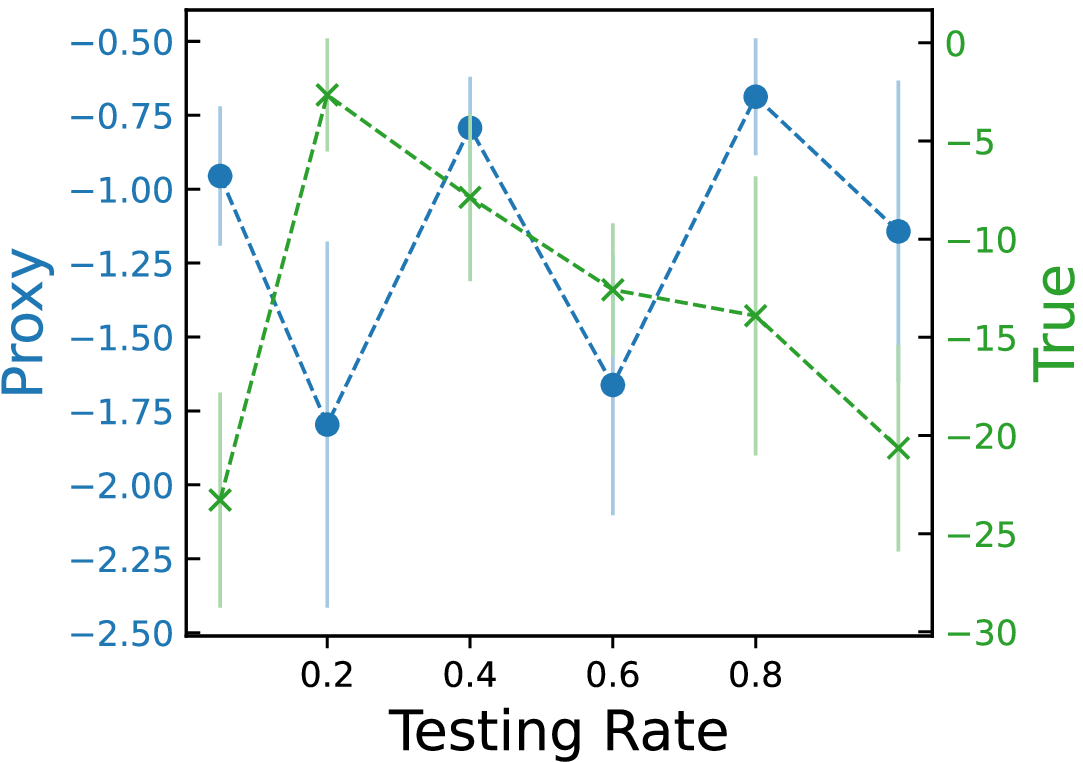}}%
    \caption{In addition to parameter count, we consider three other agent capabilities: training steps, action space resolution, and observation noise. In Figure~\ref{fig:training_steps}, an increase in the proxy reward comes at the cost of the true reward. In Figure~\ref{fig:action_noise}, increasing the granularity (from right to left) causes the agent to achieve similar proxy reward but lower true reward. In Figure~\ref{fig:obs_noise}, increasing the fidelity of observations (by increasing the random testing rate in the population) tends to decrease the true reward with no clear impact on proxy reward.}
\end{figure}

\paragraph{Model Capacity.} Our results are shown in Figure~\ref{fig:modelsize_scatter}, with additional plots included in Appendix~\ref{app:measurement}. We plot both the proxy (blue) and true (green) reward vs.~the number of parameters. 
As model size increases, the proxy reward increases but the true reward decreases. This suggests that reward designers will likely need to take greater care to specify reward functions accurately and is especially salient given the recent trends towards larger and larger models~\citep{ai100report2021}.  

The drop in true reward is sometimes quite sudden. We call these sudden shifts \emph{phase transitions}, and mark them with dashed red lines in Figure~\ref{fig:modelsize_scatter}. These quantitative trends are reflected in the qualitative behavior of the policies (Section~\ref{sec:qualitative}), which typically also shift at the phase transition. 

Model capacity is only one proxy for agent capabilities, and larger 
models do not always lead to more capable agents~\citep{andrychowicz2020matters}. To check the robustness of our results, we consider several other measures of optimization: observation fidelity, number of training steps, and action space resolution.

\paragraph{Number of training steps.} Assuming a reasonable RL algorithm and hyperparameters, agents which are trained for more steps have more optimization power. 
We vary training steps for an agent trained on the Atari environment. The true reward incentivizes staying alive for as many frames as possible while moving smoothly. The proxy reward misweights these considerations by underpenalizing the smoothness constraint. As shown in Figure~\ref{fig:training_steps}, optimizing the proxy reward for more steps harms the true reward, after an initial period where the rewards are positively correlated.  

\paragraph{Action space resolution.} Intuitively, an agent that can take more precise actions is more capable. 
For example, as technology improves, an RL car may make course corrections every millisecond instead of every second. We study action space resolution in the traffic environment by discretizing the output space of the RL agent. Specifically, under resolution level $\varepsilon$, we round the action $a \in \mathbb{R}$ output by the RL agent to the nearest multiple of $\varepsilon$ and use that as our action.  The larger the resolution level $\varepsilon$, the lower the action space resolution. Results are shown in Figure~\ref{fig:action_noise} for a fixed model size. Increasing the resolution causes the proxy reward to remain roughly constant while the true reward decreases. 


\paragraph{Observation fidelity.} Agents with access to better input sensors, like higher-resolution cameras, 
should make more informed decisions and thus have more optimization power. 
Concretely, we study this in the COVID environment, where we increase the random testing rate in the population. The proxy reward is a linear combination of the number of infections and severity of social distancing, while the true reward also factors in political cost. 
As shown in Figure~\ref{fig:obs_noise}, as the testing rate increases, the model achieves similar proxy reward at the cost of a slightly lower true reward. 







\subsection{Qualitative Effects}\label{sec:qualitative}
In the previous section, quantitative trends showed that increasing a model's optimization power often hurts performance on the true reward. We shift our focus to understanding \emph{how} this decrease happens.
In particular, we typically observe a qualitative shift in behavior 
associated with each of the phase transitions, three of which we describe below. 

%
\begin{figure}[!bp]
\centering
\subfloat[Traffic policy of smaller network] {\label{fig:t_qual1}\includegraphics[scale=0.30]{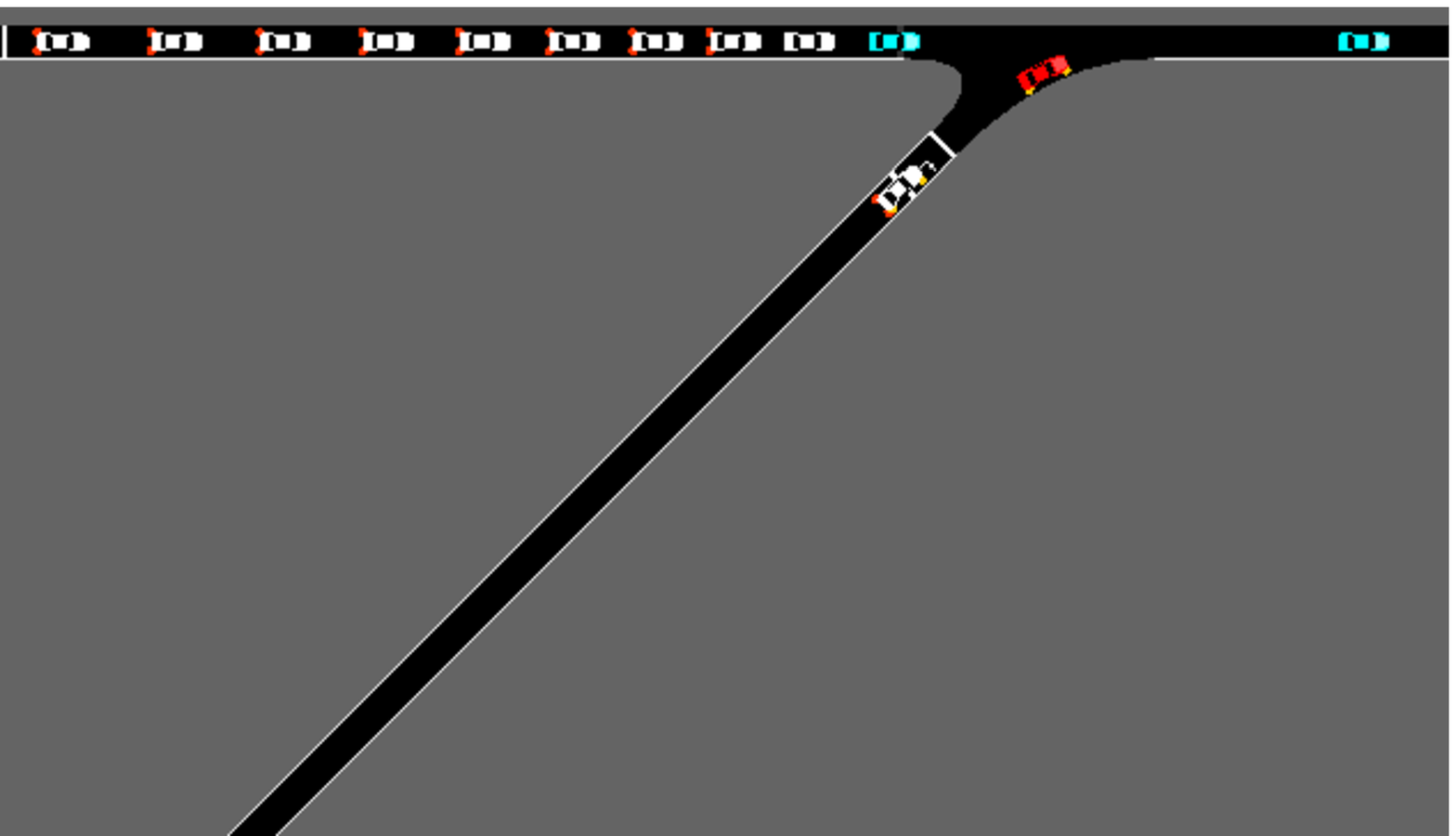}}%
\hspace{20pt}
\subfloat[Traffic policy of larger network] {\label{fig:t_qual2}\includegraphics[scale=0.30]{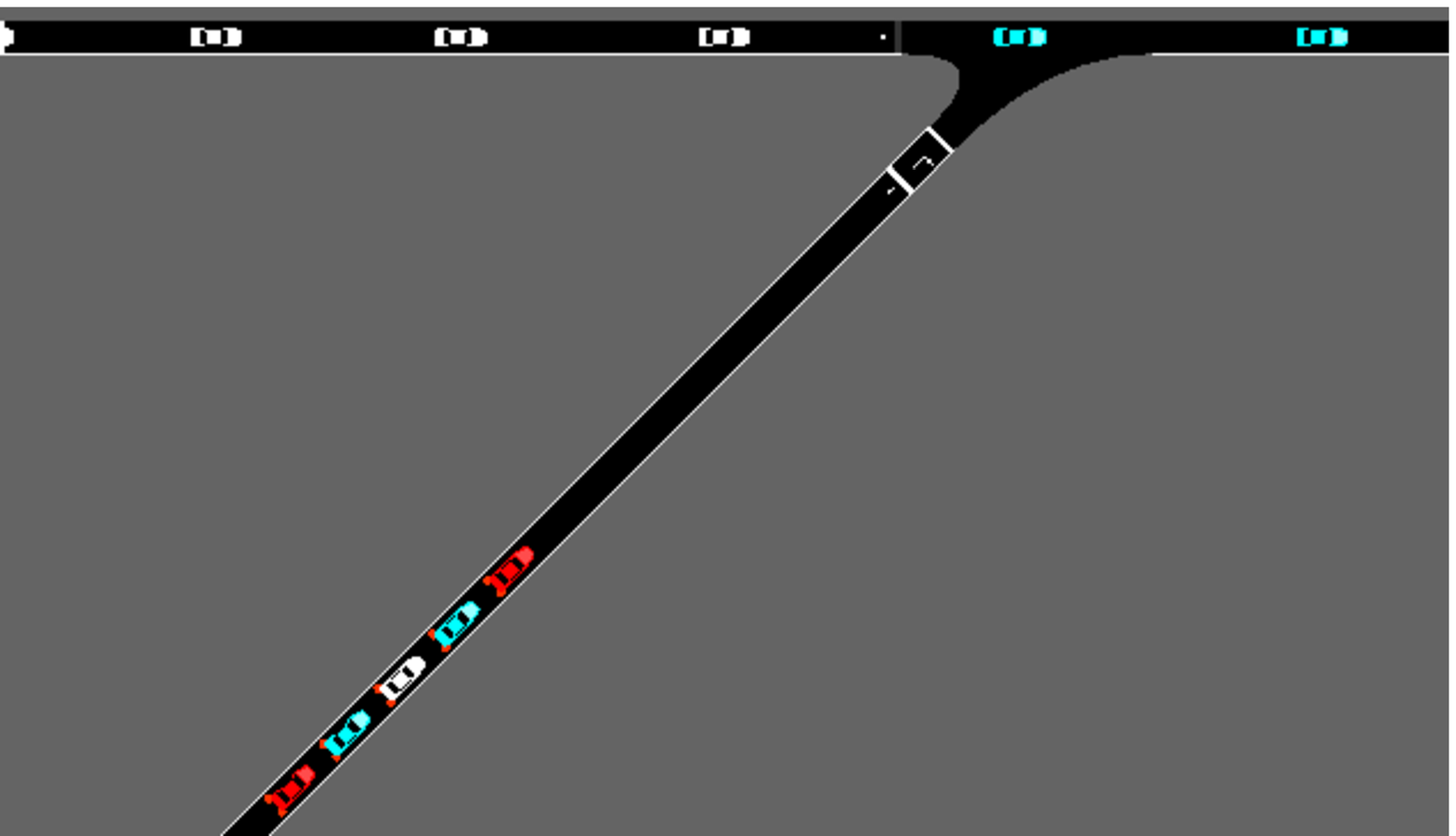}}%

\caption{The larger model prevents the AVs (in red) from moving to increase the velocity of the human cars (unobserved cars in white and observed cars in blue). However, this greatly increases the average commute per person.}
\label{fig:traffic_qualitative}
\end{figure}

\paragraph{Traffic Control.} 
We focus on the \trafficm~environment from 
Figure~\ref{fig:t_ont}, where minimizing average commute time is replaced by maximizing average velocity.
In this case, smaller policies learn to merge onto the straightaway by slightly slowing down the other vehicles (Figure~\ref{fig:t_qual1}). On the other hand, larger policy models stop the AVs to prevent them from merging at all (Figure~\ref{fig:t_qual2}). This increases the average velocity, because the vehicles on the straightaway (which greatly outnumber vehicles on the on-ramp) do not need to slow down for merging traffic. However, it significantly increases the average commute time, as the passengers in the AV remain stuck. 

\begin{figure}[!tbp]
\centering
\includegraphics[scale=0.55]{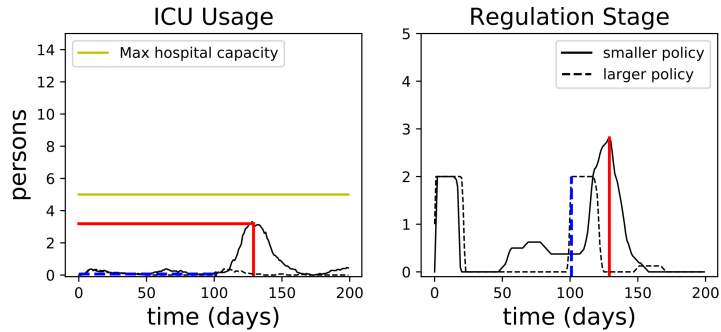}
\caption{For COVID, ICU usage is a proxy for public health and regulation stage is a proxy for economic health. The blue line indicates the maximum stage (right) enforced by the larger policy and the corresponding ICU level (left) at that stage. The red line is the equivalent for the smaller policy. Because the larger policy enforces regulations much sooner than the smaller policy, it maintains both low ICU usage and low regulation stage. However, the larger policy is politically unfavorable: regulations are high even though public signs of infection, such as ICU usage, are low.} 
\label{fig:pandemic_qualitative}
\end{figure}

\paragraph{COVID Response.}
Suppose the RL agent optimizes solely for the public and economic health of a society, without factoring politics into its decision-making. This behavior is shown in Figure~\ref{fig:pandemic_qualitative}. The larger model chooses to increase the severity of social distancing restrictions earlier than the smaller model. As a result, larger models are able to maintain low average levels of both ICU usage (a proxy for public health) and social distancing restrictions (a proxy for economic health). These preemptive regulations may however be politically costly, as enforcing restrictions without clear signs of infection may foment public unrest~\citep{boettke2021covidpolitical}.

\paragraph{Atari Riverraid.} 
We create an ontological misspecification by rewarding the plane for staying alive as long as possible while shooting as little as possible: a ``pacifist run''. We then measure the game score as the true reward.  
We find that agents with more parameters typically maneuver more adeptly. Such agents shoot less frequently, but survive for much longer, acquiring points (true reward) due to passing checkpoints. In this case, therefore, the proxy and true rewards are well-aligned so that reward hacking does not emerge as capabilities increase. 

We did, however, find that some of the agents exploited a bug in the simulator that halts the plane at the beginning of the level. The simulator advances but the plane itself does not move, thereby achieving high pacifist reward. 


\paragraph{Glucose Monitoring.} 
Consider an RL agent that optimizes solely for a patient's health, without considering the economic costs of its treatment plans. In this case, the proxy reward is based off of a glycemic risk measure, which reflects the likelihood that a patient will suffer an acute hypoglycemic episode, developed by the medical community~\citep{kovatchev2000diabetes}. 

However, a less economically-privileged patient may opt for the treatment plan with the least expected cost~\citep{herkert2019insulincost,fralick2019insulin}, not the one with the least amount of risk. From this patient's perspective, the true reward is the expected cost of the treatment plan, which includes the expected cost of hospital visits and the cost of administering the insulin. 

Although larger model treatments reduce hypoglycemic risk more smaller model treatments, they administer more insulin. Based on the average cost of an ER visit for a hypogylcemic episode ($\$1350$ from~\citet{bronstone2016hypoglycemic_cost}) and the average cost of a unit of insulin ($\$0.32$ from~\citet{lee2020insulin}), we find that it is actually more expensive to pursue the larger model's treatment. 


\subsection{Quantitative Effects vs Proxy-True Reward Correlation}\label{sec:correlation}

We saw in Sections~\ref{sec:quantitative} and~\ref{sec:qualitative} that agents often pursue proxy rewards at the cost of the true reward. Perhaps this only occurs because the proxy is greatly misspecified, i.e., the proxy and true reward are weakly or negatively correlated. If this were the case, then reward hacking may pose less of a threat. To investigate this intuition, we plot the correlation between the proxy and true rewards. 

The correlation is determined by the state distribution of a given policy, so we consider two types of state distributions. Specifically, for a given model size, we obtain two checkpoints: one that achieves the highest proxy reward during training and one from early in training (less than $1\%$ of training complete). We call the former the ``trained checkpoint" and the latter the ``early checkpoint". 

\begin{figure}[!tp]
\centering
\subfloat[\label{fig:corr_space_1}\textit{\trafficm} - Space] {\includegraphics[scale=0.45]{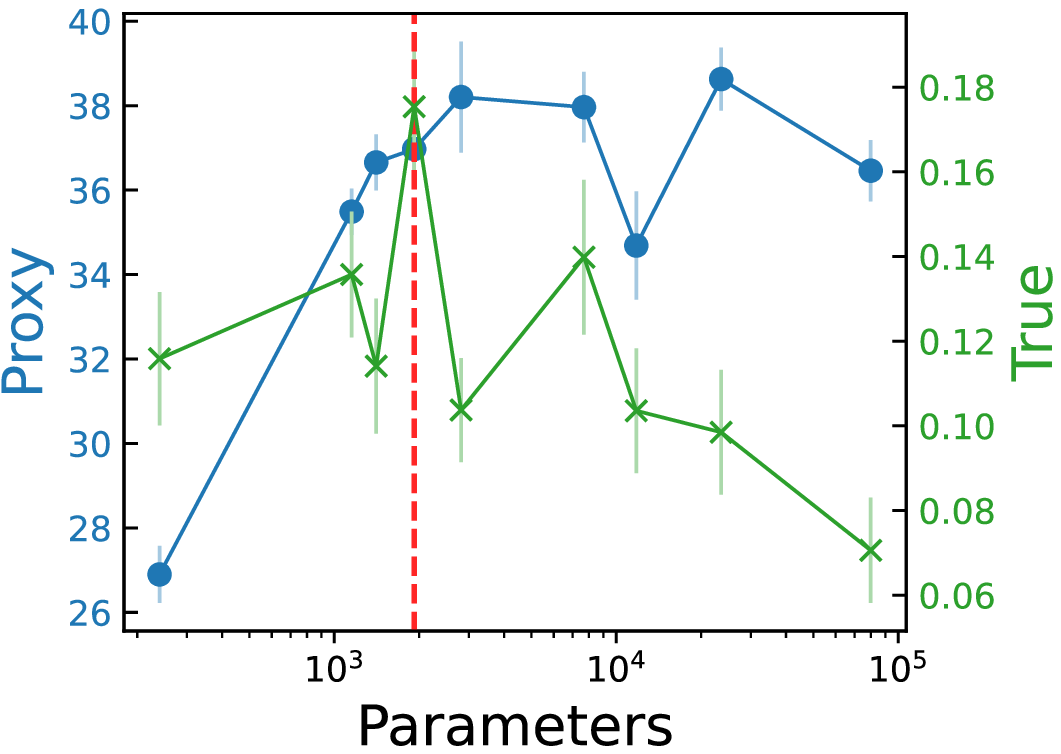}}%
\hspace{20pt}
\subfloat[\label{fig:corr_space_2}Correlation for Figure~\ref{fig:corr_space_1}] {\includegraphics[scale=0.45]{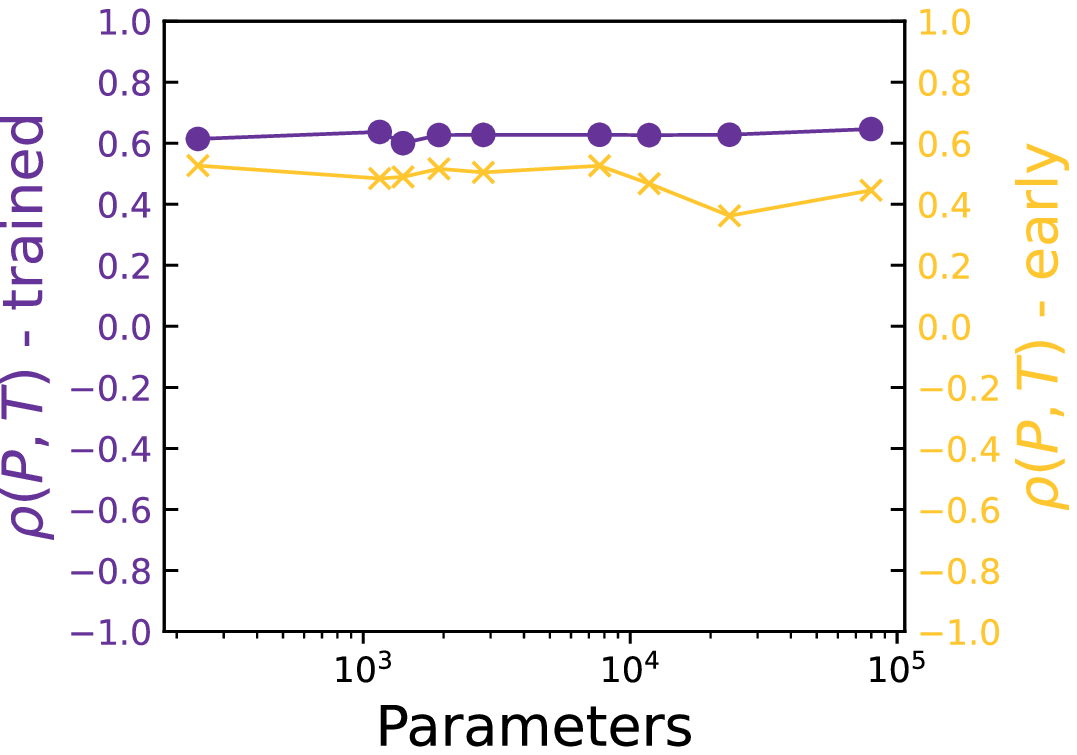}}%
\caption{Correlations between the proxy and true rewards, along with the reward hacking induced. In Figure~\ref{fig:corr_space_1}, we plot the \textcolor{bluetext}{proxy reward} with ``$\bullet$" and the \textcolor{greentext}{true reward} with ``$\times$". In Figure~\ref{fig:corr_space_2}, we plot the \textcolor{purpletext}{trained checkpoint correlation} and the \textcolor{yellowtext}{early checkpoint correlation}.}
\label{fig:corr_main}
\end{figure}

For a given model checkpoint, we calculate the Pearson correlation $\rho$ between the proxy reward $P$ and true reward $T$ using 30 trajectory rollouts. 
Reward hacking occurs even though there is significant positive correlation between the true and proxy rewards (see Figure~\ref{fig:corr_main}). The correlation is lower for the trained model than for the early model, but still high. Further figures are shown in Appendix~\ref{app:measurement_correlation}. Among the four environments tested, only the \trafficm~environment with ontological misspecification had negative Pearson correlation.

\section{Polynomaly: Mitigating reward misspecification}\label{sec:detection}
In Section~\ref{sec:measurement}, we saw that reward hacking often leads to phase transitions in agent behaviour. Furthermore, in applications like traffic control or COVID response, the true reward may be observed only sporadically or not at all. 
Blindly optimizing the proxy in these cases can lead to catastrophic failure~\citep{zhuang2020misaligned, taylor2016impossibility}. 

This raises an important question: Without the true reward signal, how can we mitigate misalignment? We operationalize this as an anomaly detection task: the detector should flag instances of misalignment, thus preventing catastrophic rollouts. To aid the detector, we provide it with a \emph{trusted policy}: one verified by humans to have acceptable (but not maximal) reward. Our resulting benchmark, \textsc{Polynomaly}, is described below.




\subsection{Problem Setup}
We train a collection of policies by varying model size on the traffic and COVID environments. For each policy, we estimate the policy's true reward by averaging over $5$ to $32$ rollouts. One author labeled each policy as {acceptable}, {problematic}, or {ambiguous} based on its true reward score relative to that of other policies. We include only policies that received a non-ambiguous label.

For both environments, we provide a small-to-medium sized model as the trusted policy model, as Section~\ref{sec:quantitative} empirically illustrates that smaller models achieve reasonable true reward without exhibiting reward hacking. Given the trusted model and a collection of policies, the anomaly detector's task is to predict the binary label of ``acceptable" or ``problematic" for each policy.

Table~\ref{tab:bench-stats} in Appendix~\ref{app:benchmark_statistics} summarizes our benchmark. The trusted policy size is a list of the hidden unit widths of the trusted policy network (not including feature mappings).



\subsection{Evaluation}
We propose two evaluation metrics for measuring the performance of our anomaly detectors. 
\begin{itemize}[leftmargin=*]
    \item  \textit{Area Under the Receiver Operating Characteristic (AUROC)}. The AUROC measures the probability that a detector will assign a random anomaly a higher score than a random non-anomalous policy~\citep{davis2006auroc}. Higher AUROCs indicate stronger detectors. 
    
    \item \textit{Max F-1 score}. The F-1 score is the harmonic mean of the precision and the recall, so detectors with a high F-1 score have both low false positives and high true negatives. We calculate the max F-1 score by taking the maximum F-1 score over all possible thresholds for the detector.
\end{itemize}


\subsection{Baselines}\label{subsec:baseline_detector}
In addition to the benchmark datasets described above, we provide baseline anomaly detectors based on estimating distances between policies. We estimate the distance between the trusted policy and the unknown policy based on either the Jensen-Shannon divergence (JSD) or the Hellinger distance. Specifically, we use rollouts to generate empirical action distributions. We compute the distance between these action distributions at each step of the rollout, then aggregate across steps by taking either the mean or the range. For full details, see Appendix~\ref{app:roc}. Table~\ref{tab:detectors} reports the AUROC and F-1 scores of several such detectors. We provide full ROC curves in Appendix~\ref{app:roc}.

\begin{table*}[htbp]
\makebox[\textwidth][c]{
    \begin{tabular}{ccccccc}
\toprule
  \makebox[6em]{\textbf{Baseline Detectors}}  & \multicolumn{2}{c}{Mean Jensen-Shannon} & \multicolumn{2}{c}{Mean Hellinger} &\multicolumn{2}{c}{Range Hellinger} \\
    \cmidrule(lr){1-1}\cmidrule(lr){2-3}\cmidrule(lr){4-5}\cmidrule(lr){6-7}
Env. - Misspecification    & AUROC & Max F-1 & AUROC & Max F-1 & AUROC & Max F-1 \\ \midrule
\trafficm~- misweighting  & $81.0\%$ & $0.824$ & $81.0\%$ & $0.824$ & $76.2\%$& $0.824$\\
\trafficm~- scope  & $74.6\%$ & $0.818$& $74.6\%$ & $0.818$ & $57.1\%$ & $0.720$\\
\trafficm~- ontological  & $52.7\%$ & $0.583$ & $55.4\%$& $0.646$& $71.4\%$ & $0.842$ \\
\trafficb~- misweighting & $88.9\%$ & $0.900$ & $88.9\%$& $0.900$&$74.1\%$& $0.857$\\
COVID - ontological & $45.2\%$& $0.706$  & $59.5\%$& $0.750$ & $88.1\%$& $0.923$\\
\bottomrule
\end{tabular}%
}

\caption{Performance of detectors on different subtasks. Each detector has at least one subtask with AUROC under 60\%, indicating poor performance.}
\label{tab:detectors}%
\end{table*}%
We observe that different detectors are better for different tasks, suggesting that future detectors could do better than any of our baselines. 
Our benchmark and baseline provides a 
starting point for further research on mitigating reward hacking.

\section{Discussion}


In this work, we designed a diverse set of environments and proxy rewards, uncovered several instances of phase transitions, and proposed an anomaly detection task to help mitigate these transitions. Our results raise two questions: How can we not only detect phase transitions, but prevent them in the first place? And how should phase transitions shape our approach to safe ML?

On preventing phase transitions, anomaly detection already offers one path forward. Once we can detect anomalies, we can potentially prevent them, by using the detector 
to purge the unwanted behavior (e.g.~by including it in the training objective). Similar policy shaping has recently been used to 
make RL agents more ethical \citep{hendrycks2021would}. However, since the anomaly detectors will be optimized against by the RL policy, they need to be adversarially robust \citep{goodfellow2014explaining}. This motivates further work on adversarial robustness and adversarial anomaly detection. Another possible direction is optimizing policies against a distribution of rewards~\citep{brown2019bayesianrex,javed2021pgbroil}, which may prevent over-fitting to a given set of metrics.

Regarding safe ML, several recent papers propose extrapolating empirical trends to forecast future ML capabilities \citep{kaplan2020scaling,hernandez2021scaling,droppo2021scaling}, partly to avoid unforeseen consequences from ML. While we support this work, our results show that trend extrapolation alone is not enough to ensure the safety of ML systems. To complement trend extrapolation, we need better interpretability methods to identify emergent model behaviors early on, before they dominate performance \citep{olah2018building}. ML researchers should also familiarize themselves with emergent behavior in self-organizing systems \citep{yates2012self}, which often exhibit similar phase transitions \citep{anderson1972more}. Indeed, the ubiquity of phase transitions throughout science suggests that ML researchers should continue to expect surprises--and should therefore prepare for them.

\newpage
\section*{Acknowledgements}
We are thankful to Dan Hendrycks and Adam Gleave for helpful discussions about experiments and to Cassidy Laidlaw and Dan Hendrycks for providing valuable feedback on the writing. KB was supported by a JP Morgan AI Fellowship. JS was supported by NSF Award 2031985 and by Open Philanthropy.

\newpage
\newpage
\bibliography{ref}

\begin{thebibliography}{54}
\providecommand{\natexlab}[1]{#1}
\providecommand{\url}[1]{\texttt{#1}}
\expandafter\ifx\csname urlstyle\endcsname\relax
  \providecommand{\doi}[1]{doi: #1}\else
  \providecommand{\doi}{doi: \begingroup \urlstyle{rm}\Url}\fi

\bibitem[Anderson(1972)]{anderson1972more}
Philip~W Anderson.
\newblock More is different.
\newblock \emph{Science}, 177\penalty0 (4047):\penalty0 393--396, 1972.

\bibitem[Andrychowicz et~al.(2020)Andrychowicz, Raichuk, Sta{\'n}czyk, Orsini,
  Girgin, Marinier, Hussenot, Geist, Pietquin, and
  Michalski]{andrychowicz2020matters}
Marcin Andrychowicz, Anton Raichuk, Piotr Sta{\'n}czyk, Manu Orsini, Sertan
  Girgin, Raphael Marinier, L{\'e}onard Hussenot, Matthieu Geist, Olivier
  Pietquin, and Marcin Michalski.
\newblock What matters in on-policy reinforcement learning? {A} large-scale
  empirical study.
\newblock \emph{arXiv preprint arXiv:2006.05990}, 2020.

\bibitem[Baker et~al.(2020)Baker, Kanitscheider, Markov, Wu, Powell, McGrew,
  and Mordatch]{baker2020emergent}
Bowen Baker, Ingmar Kanitscheider, Todor Markov, Yi~Wu, Glenn Powell, Bob
  McGrew, and Igor Mordatch.
\newblock Emergent tool use from multi-agent autocurricula.
\newblock In \emph{International Conference on Learning Representations}, 2020.

\bibitem[Boettke \& Powell(2021)Boettke and Powell]{boettke2021covidpolitical}
Peter Boettke and Benjamin Powell.
\newblock The political economy of the covid-19 pandemic.
\newblock \emph{Southern Economic Journal}, 87\penalty0 (4):\penalty0
  1090--1106, 2021.

\bibitem[Bommasani et~al.(2021)]{bommansani2021foundational}
Rishi Bommasani et~al.
\newblock On the opportunities and risks of foundation models.
\newblock \emph{arXiv preprint arXiv:2108.07258}, 2021.

\bibitem[Brockman et~al.(2016)Brockman, Cheung, Pettersson, Schneider,
  Schulman, Tang, and Zaremba]{brockman2016gym}
Greg Brockman, Vicki Cheung, Ludwig Pettersson, Jonas Schneider, John Schulman,
  Jie Tang, and Wojciech Zaremba.
\newblock Openai gym, 2016.

\bibitem[Bronstone \& Graham(2016)Bronstone and
  Graham]{bronstone2016hypoglycemic_cost}
Amy Bronstone and Claudia Graham.
\newblock The potential cost implications of averting severe hypoglycemic
  events requiring hospitalization in high-risk adults with type 1 diabetes
  using real-time continuous glucose monitoring.
\newblock \emph{Journal of Diabetes Science and Technology}, 10, 2016.

\bibitem[Brown et~al.(2020)Brown, Coleman, Srinivasan, and
  Niekum]{brown2019bayesianrex}
Daniel Brown, Russell Coleman, Ravi Srinivasan, and Scott Niekum.
\newblock Safe imitation learning via fast {B}ayesian reward inference from
  preferences.
\newblock In \emph{Proceedings of the 37th International Conference on Machine
  Learning}, 2020.

\bibitem[Christiano et~al.(2017)Christiano, Leike, Brown, Martic, Legg, and
  Amodei]{christiano2017preflearning}
Paul~F Christiano, Jan Leike, Tom Brown, Miljan Martic, Shane Legg, and Dario
  Amodei.
\newblock Deep reinforcement learning from human preferences.
\newblock In \emph{Advances in Neural Information Processing Systems}, 2017.

\bibitem[Davis \& Goadrich(2006)Davis and Goadrich]{davis2006auroc}
Jesse Davis and Mark Goadrich.
\newblock The relationship between precision-recall and roc curves.
\newblock In \emph{International Conference on Machine Learning}, 2006.

\bibitem[Droppo \& Elibol(2021)Droppo and Elibol]{droppo2021scaling}
Jasha Droppo and Oguz Elibol.
\newblock Scaling laws for acoustic models.
\newblock \emph{arXiv preprint arXiv:2106.09488}, 2021.

\bibitem[Espeholt et~al.(2018)Espeholt, Soyer, Munos, Simonyan, Mnih, Ward,
  Doron, Firoiu, Harley, Dunning, Legg, and Kavukcuoglu]{espeholt2018impala}
Lasse Espeholt, Hubert Soyer, Remi Munos, Karen Simonyan, Volodymyr Mnih, Tom
  Ward, Yotam Doron, Vlad Firoiu, Tim Harley, Iain~Robert Dunning, Shane Legg,
  and Koray Kavukcuoglu.
\newblock Impala: Scalable distributed deep-rl with importance weighted
  actor-learner architectures.
\newblock 2018.

\bibitem[Everitt et~al.(2017)Everitt, Krakovna, Orseau, and
  Legg]{everitt2017corruptedreward}
Tom Everitt, Victoria Krakovna, Laurent Orseau, and Shane Legg.
\newblock Reinforcement learning with a corrupted reward channel.
\newblock In \emph{International Joint Conference on Artificial Intelligence},
  2017.

\bibitem[Fox et~al.(2020)Fox, Lee, Pop-Busui, and Wiens]{fox2020bgp}
Ian Fox, Joyce Lee, Rodica Pop-Busui, and Jenna Wiens.
\newblock Deep reinforcement learning for closed-loop blood glucose control.
\newblock In \emph{Machine Learning for Healthcare Conference}, 2020.

\bibitem[Fralick \& Kesselheim(2019)Fralick and Kesselheim]{fralick2019insulin}
M.~Fralick and A.~S. Kesselheim.
\newblock {{T}he {U}.{S}. Insulin Crisis - {R}ationing a Lifesaving Medication
  Discovered in the 1920s}.
\newblock \emph{New England Journal of Medicine}, 381\penalty0 (19):\penalty0
  1793--1795, 2019.

\bibitem[Goodfellow et~al.(2014)Goodfellow, Shlens, and
  Szegedy]{goodfellow2014explaining}
Ian~J Goodfellow, Jonathon Shlens, and Christian Szegedy.
\newblock Explaining and harnessing adversarial examples.
\newblock \emph{arXiv preprint arXiv:1412.6572}, 2014.

\bibitem[Haarnoja et~al.(2018)Haarnoja, Zhou, Abbeel, and
  Levine]{haarnoja2018sac}
Tuomas Haarnoja, Aurick Zhou, Pieter Abbeel, and Sergey Levine.
\newblock Soft actor-critic: Off-policy maximum entropy deep reinforcement
  learning with a stochastic actor.
\newblock In \emph{International conference on machine learning}, 2018.

\bibitem[Hadfield-Menell et~al.(2017)Hadfield-Menell, Milli, Abbeel, Russell,
  and Dragan]{hadfield2017ird}
Dylan Hadfield-Menell, Smitha Milli, Pieter Abbeel, Stuart~J Russell, and Anca
  Dragan.
\newblock Inverse reward design.
\newblock In \emph{Advances in Neural Information Processing Systems}, 2017.

\bibitem[Hendrycks \& Gimpel(2017)Hendrycks and Gimpel]{Hendrycks2017ABF}
Dan Hendrycks and Kevin Gimpel.
\newblock A baseline for detecting misclassified and out-of-distribution
  examples in neural networks.
\newblock \emph{International Conference on Learning Representations}, 2017.

\bibitem[Hendrycks et~al.(2021{\natexlab{a}})Hendrycks, Carlini, Schulman, and
  Steinhardt]{Hendrycks2021UnsolvedPI}
Dan Hendrycks, Nicholas Carlini, John Schulman, and Jacob Steinhardt.
\newblock Unsolved problems in ml safety.
\newblock \emph{arXiv preprint arXiv:2109.13916}, 2021{\natexlab{a}}.

\bibitem[Hendrycks et~al.(2021{\natexlab{b}})Hendrycks, Mazeika, Zou, Patel,
  Zhu, Navarro, Song, Li, and Steinhardt]{hendrycks2021would}
Dan Hendrycks, Mantas Mazeika, Andy Zou, Sahil Patel, Christine Zhu, Jesus
  Navarro, Dawn Song, Bo~Li, and Jacob Steinhardt.
\newblock What would {J}iminy {C}ricket do? {T}owards agents that behave
  morally.
\newblock 2021{\natexlab{b}}.

\bibitem[Herkert et~al.(2019)Herkert, Vijayakumar, Luo, Schwartz, Rabin,
  DeFilippo, and Lipska]{herkert2019insulincost}
Darby Herkert, Pavithra Vijayakumar, Jing Luo, Jeremy~I. Schwartz, Tracy~L.
  Rabin, Eunice DeFilippo, and Kasia~J. Lipska.
\newblock Cost-related insulin underuse among patients with diabetes.
\newblock \emph{JAMA Internal Medicine}, 179\penalty0 (1):\penalty0 112--114,
  Jan 2019.

\bibitem[Hernandez et~al.(2021)Hernandez, Kaplan, Henighan, and
  McCandlish]{hernandez2021scaling}
Danny Hernandez, Jared Kaplan, Tom Henighan, and Sam McCandlish.
\newblock Scaling laws for transfer.
\newblock \emph{arXiv preprint arXiv:2102.01293}, 2021.

\bibitem[Hubinger et~al.(2019)Hubinger, van Merwijk, Mikulik, Skalse, and
  Garrabrant]{hubinger2019risks}
Evan Hubinger, Chris van Merwijk, Vladimir Mikulik, Joar Skalse, and Scott
  Garrabrant.
\newblock Risks from learned optimization in advanced machine learning systems.
\newblock \emph{arXiv preprint arXiv:1906.01820}, 2019.

\bibitem[Ibarz et~al.(2018)Ibarz, Leike, Pohlen, Irving, Legg, and
  Amodei]{ibarz2018humandemo}
Borja Ibarz, J.~Leike, Tobias Pohlen, Geoffrey Irving, S.~Legg, and Dario
  Amodei.
\newblock Reward learning from human preferences and demonstrations in {A}tari.
\newblock In \emph{Advances in Neural Information Processing Systems}, 2018.

\bibitem[Javed et~al.(2021)Javed, Brown, Sharma, Zhu, Balakrishna, Petrik,
  Dragan, and Goldberg]{javed2021pgbroil}
Zaynah Javed, Daniel~S Brown, Satvik Sharma, Jerry Zhu, Ashwin Balakrishna,
  Marek Petrik, Anca Dragan, and Ken Goldberg.
\newblock Policy gradient bayesian robust optimization for imitation learning.
\newblock In \emph{Proceedings of the 38th International Conference on Machine
  Learning}, 2021.

\bibitem[Kaplan et~al.(2020)Kaplan, McCandlish, Henighan, Brown, Chess, Child,
  Gray, Radford, Wu, and Amodei]{kaplan2020scaling}
Jared Kaplan, Sam McCandlish, Tom Henighan, Tom~B Brown, Benjamin Chess, Rewon
  Child, Scott Gray, Alec Radford, Jeffrey Wu, and Dario Amodei.
\newblock Scaling laws for neural language models.
\newblock \emph{arXiv preprint arXiv:2001.08361}, 2020.

\bibitem[{Knox} et~al.(2021){Knox}, {Allievi}, {Banzhaf}, {Schmitt}, and
  {Stone}]{knox2021misdesign}
W.~Bradley {Knox}, Alessandro {Allievi}, Holger {Banzhaf}, Felix {Schmitt}, and
  Peter {Stone}.
\newblock {Reward (Mis)design for Autonomous Driving}.
\newblock \emph{arXiv e-prints arXiv:2104.13906}, 2021.

\bibitem[Kober et~al.(2013)Kober, Bagnell, and Peters]{kober2013robotics}
Jens Kober, J~Andrew Bagnell, and Jan Peters.
\newblock Reinforcement learning in robotics: A survey.
\newblock \emph{The International Journal of Robotics Research}, 32\penalty0
  (11):\penalty0 1238--1274, 2013.

\bibitem[Kompella et~al.(2020)Kompella, Capobianco, Jong, Browne, Fox, Meyers,
  Wurman, and Stone]{kompella2020pandemic}
Varun Kompella, Roberto Capobianco, Stacy Jong, Jonathan Browne, Spencer Fox,
  Lauren Meyers, Peter Wurman, and Peter Stone.
\newblock Reinforcement learning for optimization of covid-19 mitigation
  policies, 2020.

\bibitem[Kovatchev et~al.(2000)Kovatchev, Straume, Cox, and
  Farhy]{kovatchev2000diabetes}
BorIs.~P. Kovatchev, Martin Straume, Daniel~J. Cox, and Leon.S Farhy.
\newblock Risk analysis of blood glucose data:a quantitative approach to
  optimizing the control of insulin dependent diabetes.
\newblock \emph{Journal of Theoretical Medicine}, 3\penalty0 (1):\penalty0
  1--10, 2000.

\bibitem[K\"{u}ttler et~al.(2019)K\"{u}ttler, Nardelli, Lavril, Selvatici,
  Sivakumar, Rockt\"{a}schel, and Grefenstette]{torchbeast2019}
Heinrich K\"{u}ttler, Nantas Nardelli, Thibaut Lavril, Marco Selvatici,
  Viswanath Sivakumar, Tim Rockt\"{a}schel, and Edward Grefenstette.
\newblock {TorchBeast: A PyTorch Platform for Distributed RL}.
\newblock \emph{arXiv preprint arXiv:1910.03552}, 2019.

\bibitem[Lee(2020)]{lee2020insulin}
Benita Lee.
\newblock How much does insulin cost? {H}ere's how 23 brands compare, Nov 2020.

\bibitem[Leike et~al.(2017)Leike, Martic, Krakovna, Ortega, Everitt, Lefrancq,
  Orseau, and Legg]{leike2017gridworld}
Jan Leike, Miljan Martic, Victoria Krakovna, Pedro~A. Ortega, Tom Everitt,
  Andrew Lefrancq, Laurent Orseau, and Shane Legg.
\newblock {AI} safety gridworlds, 2017.

\bibitem[Littman et~al.(2021)Littman, Ajunwa, Berger, Boutilier, Currie,
  Doshi-Velez, Hadfield, Horowitz, Isbell, Kitano, Levy, Lyons, Mitchell, Shah,
  Sloman, Vallor, and Walsh]{ai100report2021}
Michael~L. Littman, Ifeoma Ajunwa, Guy Berger, Craig Boutilier, Morgan Currie,
  Finale Doshi-Velez, Gillian Hadfield, Michael~C. Horowitz, Charles Isbell,
  Hiroaki Kitano, Karen Levy, Terah Lyons, Melanie Mitchell, Julie Shah, Steven
  Sloman, Shannon Vallor, and Toby Walsh.
\newblock Gathering strength, gathering storms: The one hundred year study on
  artificial intelligence ({AI100}) 2021 study panel report.
\newblock Technical report, Stanford University, Stanford, CA, 2021.

\bibitem[Lopez et~al.(2018)Lopez, Behrisch, Bieker-Walz, Erdmann,
  Fl{\"o}tter{\"o}d, Hilbrich, L{\"u}cken, Rummel, Wagner, and
  Wie{\ss}ner]{sumo2018}
Pablo~Alvarez Lopez, Michael Behrisch, Laura Bieker-Walz, Jakob Erdmann,
  Yun-Pang Fl{\"o}tter{\"o}d, Robert Hilbrich, Leonhard L{\"u}cken, Johannes
  Rummel, Peter Wagner, and Evamarie Wie{\ss}ner.
\newblock Microscopic traffic simulation using {SUMO}.
\newblock In \emph{International Conference on Intelligent Transportation
  Systems}, 2018.

\bibitem[Man et~al.(2014)Man, Micheletto, Lv, Breton, Kovatchev, and
  Cobelli]{man2014diabetes}
Chiara~Dalla Man, Francesco Micheletto, Dayu Lv, Marc Breton, Boris Kovatchev,
  and Claudio Cobelli.
\newblock The {UVA/PADOVA} type 1 diabetes simulator: New features.
\newblock \emph{Journal of Diabetes Science and Technology}, 8\penalty0
  (1):\penalty0 26--34, Jan 2014.

\bibitem[Olah et~al.(2018)Olah, Satyanarayan, Johnson, Carter, Schubert, Ye,
  and Mordvintsev]{olah2018building}
Chris Olah, Arvind Satyanarayan, Ian Johnson, Shan Carter, Ludwig Schubert,
  Katherine Ye, and Alexander Mordvintsev.
\newblock The building blocks of interpretability.
\newblock \emph{Distill}, 3\penalty0 (3):\penalty0 e10, 2018.

\bibitem[Paulus et~al.(2018)Paulus, Xiong, and
  Socher]{paulus2018drl_summarization}
Romain Paulus, Caiming Xiong, and Richard Socher.
\newblock A deep reinforced model for abstractive summarization.
\newblock In \emph{International Conference on Learning Representations}, 2018.

\bibitem[Ribeiro et~al.(2020)Ribeiro, Ottoni, West, Almeida, and
  Meira]{times2019brazil}
Manoel~Horta Ribeiro, Raphael Ottoni, Robert West, Virg\'{\i}lio A.~F. Almeida,
  and Wagner Meira.
\newblock Auditing radicalization pathways on youtube.
\newblock In \emph{Conference on Fairness, Accountability, and Transparency},
  New York, NY, USA, 2020.

\bibitem[Russell(2019)]{russell2019human}
Stuart Russell.
\newblock \emph{Human Compatible: Artificial Intelligence and the Problem of
  Control}.
\newblock Penguin, 2019.

\bibitem[Schulman et~al.(2017)Schulman, Wolski, Dhariwal, Radford, and
  Klimov]{schulman2017ppo}
John Schulman, Filip Wolski, Prafulla Dhariwal, Alec Radford, and Oleg Klimov.
\newblock Proximal policy optimization algorithms.
\newblock \emph{arXiv preprint arXiv:1707.06347}, 2017.

\bibitem[Stiennon et~al.(2020)Stiennon, Ouyang, Wu, Ziegler, Lowe, Voss,
  Radford, Amodei, and Christiano]{stiennon2020learning}
Nisan Stiennon, Long Ouyang, Jeff Wu, Daniel~M Ziegler, Ryan Lowe, Chelsea
  Voss, Alec Radford, Dario Amodei, and Paul Christiano.
\newblock Learning to summarize from human feedback.
\newblock \emph{arXiv preprint arXiv:2009.01325}, 2020.

\bibitem[Stray(2020)]{stray2020recommendersystem}
Jonathan Stray.
\newblock Aligning ai optimization to community well-being.
\newblock \emph{International Journal of Community Well-Being}, 3\penalty0
  (4):\penalty0 443--463, Dec 2020.

\bibitem[Tack et~al.(2020)Tack, Mo, Jeong, and Shin]{Tack2020CSIND}
Jihoon Tack, Sangwoo Mo, Jongheon Jeong, and Jinwoo Shin.
\newblock Csi: Novelty detection via contrastive learning on distributionally
  shifted instances.
\newblock \emph{Advances in Neural Information Processing Systems}, 2020.

\bibitem[Taylor(2016)]{taylor2016impossibility}
Jessica Taylor.
\newblock Quantilizers: A safer alternative to maximizers for limited
  optimization.
\newblock In \emph{AAAI Workshop: AI, Ethics, and Society}, 2016.

\bibitem[Toromanoff et~al.(2019)Toromanoff, Wirbel, and
  Moutarde]{toromanoff2019drl}
Marin Toromanoff, Emilie Wirbel, and Fabien Moutarde.
\newblock Is deep reinforcement learning really superhuman on {A}tari?
  {L}eveling the playing field, 2019.

\bibitem[Treiber et~al.(2000)Treiber, Hennecke, and Helbing]{treiber2000idm}
Martin Treiber, Ansgar Hennecke, and Dirk Helbing.
\newblock Congested traffic states in empirical observations and microscopic
  simulations.
\newblock \emph{Physical review E}, 62\penalty0 (2):\penalty0 1805, 2000.

\bibitem[{Trott} et~al.(2021){Trott}, {Srinivasa}, {van der Wal}, {Haneuse},
  and {Zheng}]{trott2021aieconomist}
Alexander {Trott}, Sunil {Srinivasa}, Douwe {van der Wal}, Sebastien {Haneuse},
  and Stephan {Zheng}.
\newblock {Building a Foundation for Data-Driven, Interpretable, and Robust
  Policy Design using the AI Economist}.
\newblock \emph{arXiv preprint arXiv:2108.02904}, 2021.

\bibitem[Vinitsky et~al.(2018)Vinitsky, Kreidieh, Flem, Kheterpal, Jang, Wu,
  Wu, Liaw, Liang, and Bayen]{vinitsky2018flowbench}
Eugene Vinitsky, Aboudy Kreidieh, Luc~Le Flem, Nishant Kheterpal, Kathy Jang,
  Cathy Wu, Fangyu Wu, Richard Liaw, Eric Liang, and Alexandre~M. Bayen.
\newblock Benchmarks for reinforcement learning in mixed-autonomy traffic.
\newblock In \emph{Conference on Robot Learning}, 2018.

\bibitem[Wu et~al.(2021)Wu, Kreidieh, Parvate, Vinitsky, and Bayen]{wu2017flow}
Cathy Wu, Abdul~Rahman Kreidieh, Kanaad Parvate, Eugene Vinitsky, and
  Alexandre~M. Bayen.
\newblock Flow: A modular learning framework for mixed autonomy traffic.
\newblock \emph{IEEE Transactions on Robotics}, 2021.

\bibitem[Yates(2012)]{yates2012self}
F~Eugene Yates.
\newblock \emph{Self-organizing systems: The emergence of order}.
\newblock Springer Science \& Business Media, 2012.

\bibitem[Yu et~al.(2019)Yu, Liu, and Nemati]{yu2019rlhealth}
Chao Yu, Jiming Liu, and Shamim Nemati.
\newblock Reinforcement learning in healthcare: A survey.
\newblock \emph{arXiv preprint arXiv:1908.08796}, 2019.

\bibitem[Zhuang \& Hadfield-Menell(2020)Zhuang and
  Hadfield-Menell]{zhuang2020misaligned}
Simon Zhuang and Dylan Hadfield-Menell.
\newblock Consequences of misaligned {AI}.
\newblock In \emph{Advances in Neural Information Processing Systems}, 2020.

\end{thebibliography}
\bibliographystyle{iclr2022_conference}
\newpage
\appendix
\section{Mapping The Effects of Reward Misspecification}\label{app:measurement}

\begin{figure}[!h]
\centering
\subfloat[\label{fig:t_mis_merge}\textit{traffic\_merge} - Misweighting] {\includegraphics[scale=0.60]{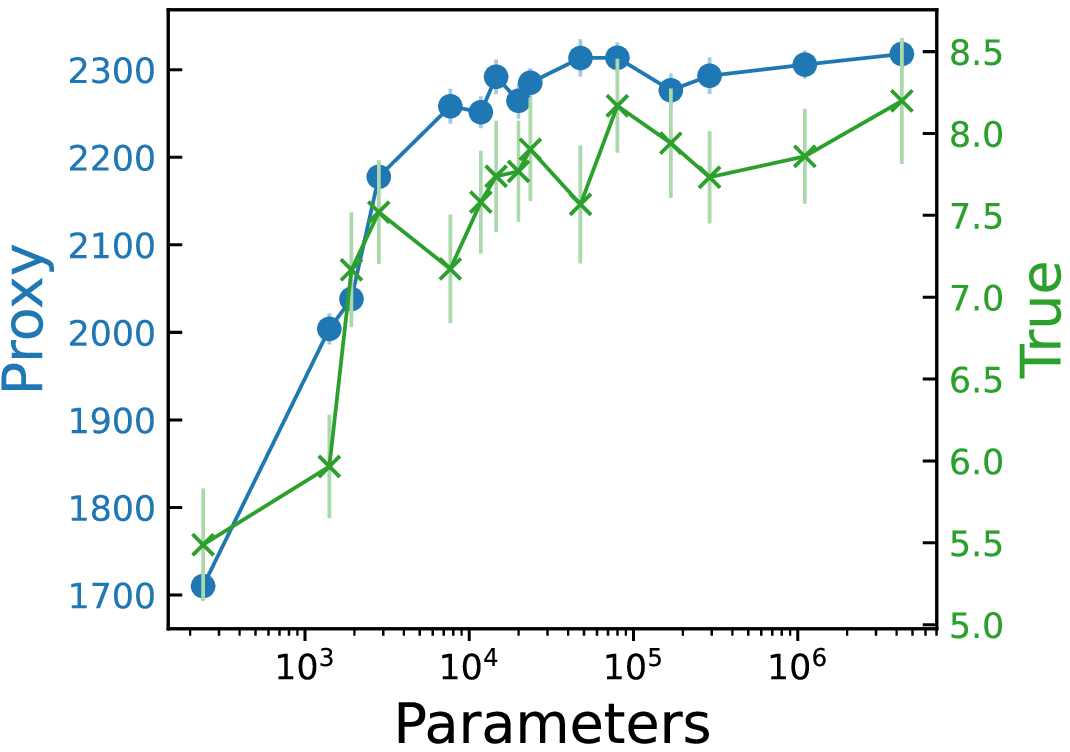}}%
\hfill
\subfloat[\label{fig:t_ont_bottle}\textit{traffic\_bottle} - Misweighting] {\includegraphics[scale=0.60]{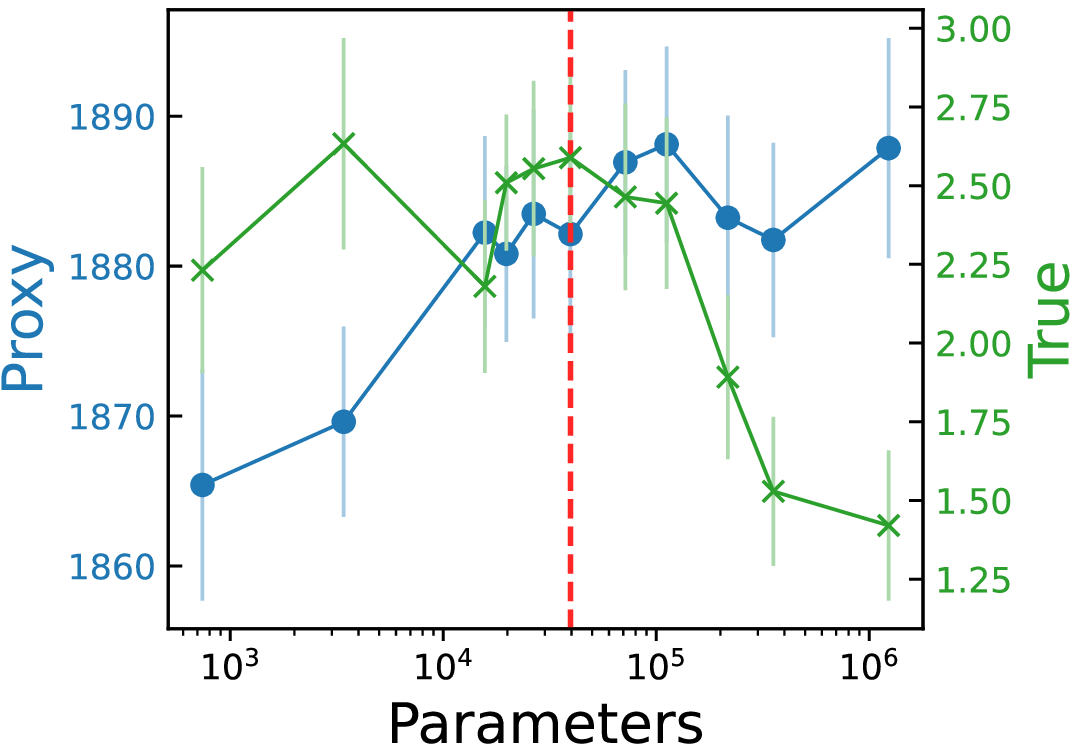}}%
\vfill
\subfloat[\label{fig:t_space_merge}\textit{traffic\_merge} - Space] {\includegraphics[scale=0.60]{figures/traffic/space_max_proxy_single.eps}}%
\hfill
\subfloat[\label{fig:p_mis}COVID - Misweighting] {\includegraphics[scale=0.60]{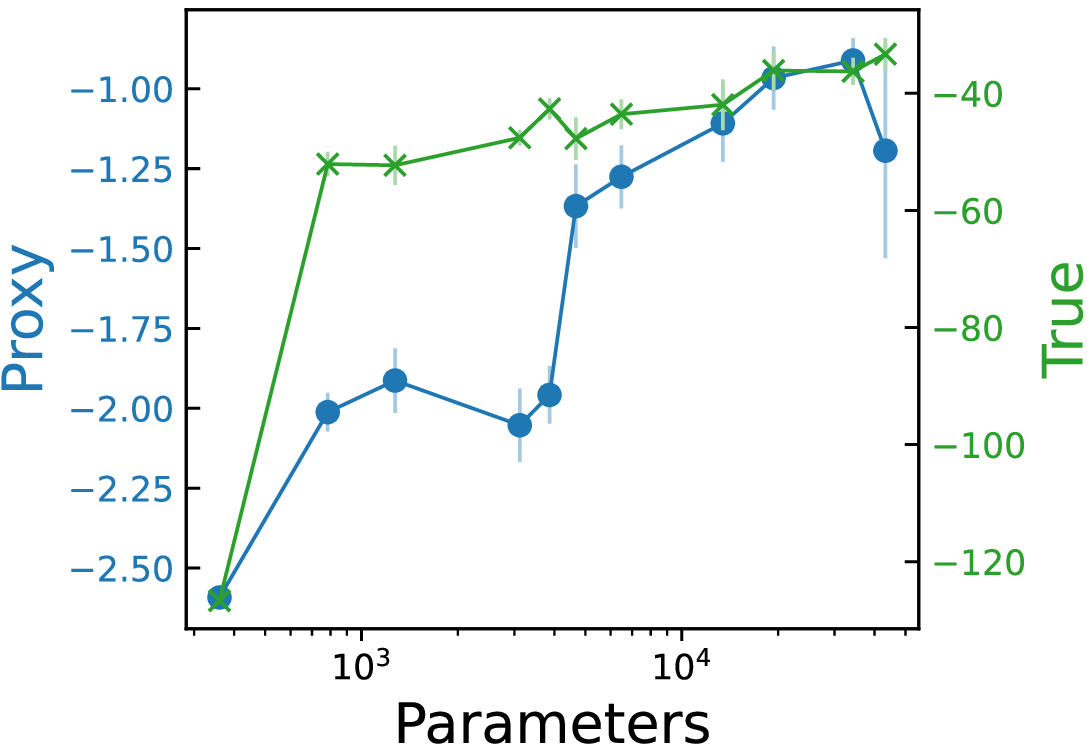}}%
\vfill
\subfloat[\label{fig:a_ont}Atari - Ontological] {\includegraphics[scale=0.60]{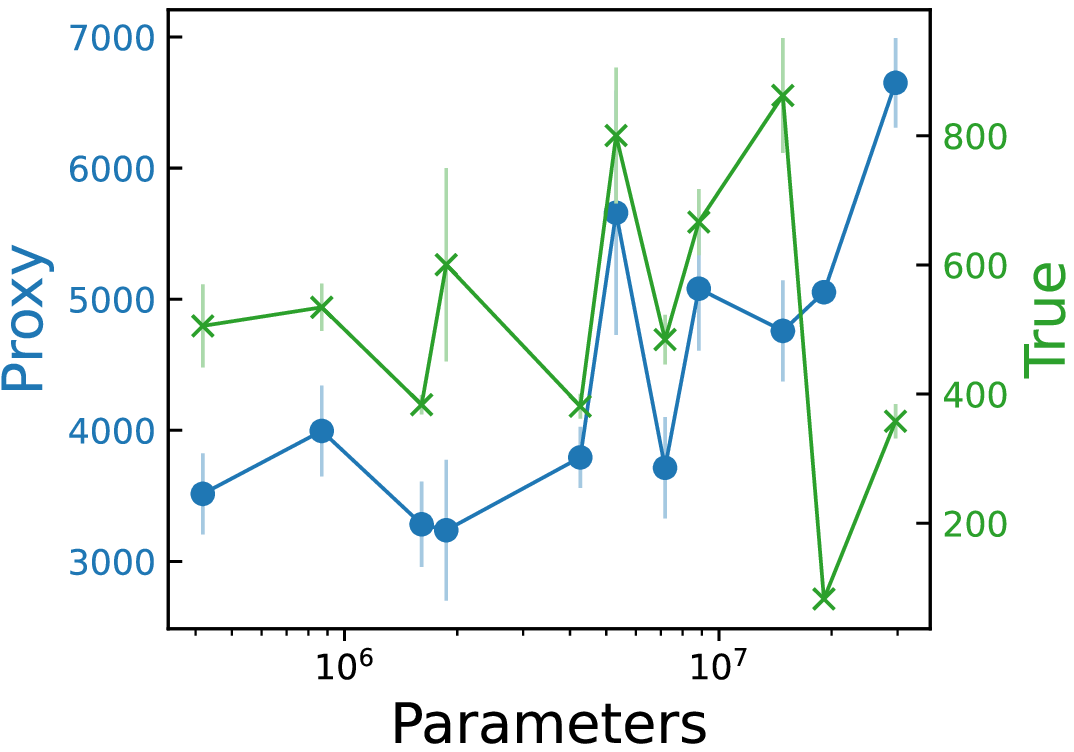}}%
\hfill
\subfloat[\label{fig:a_mis}Atari - Misweighting] {\includegraphics[scale=0.60]{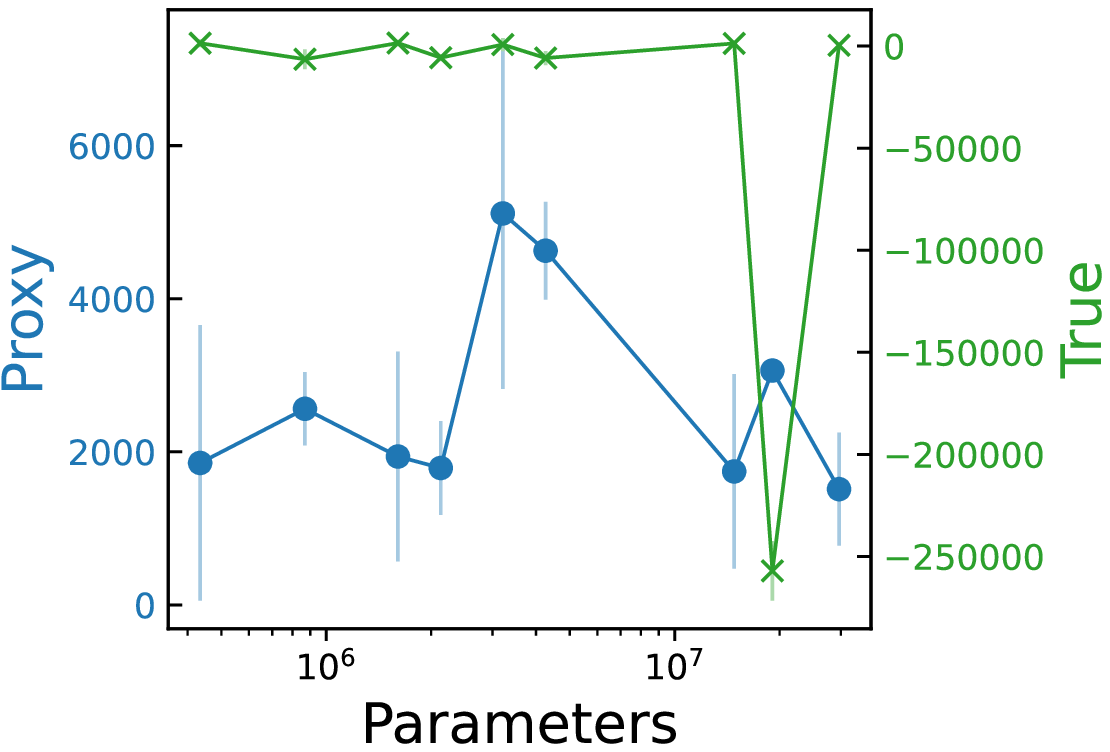}}%

\caption{Additional model size scatter plots. Observe that not all misspecifications cause misalignment. We plot the \textcolor{bluetext}{proxy reward} with ``$\bullet$" and the \textcolor{greentext}{true reward} with ``$\times$". The proxy reward is measured on the left-hand side of each figure and the true reward is measured on the right hand side of each figure.}
\label{fig:modelsize_scatter_appendix}
\end{figure}
\subsection{Effect of Model Size}\label{app:modelsize_plots}
We plot the proxy and true reward vs. model size in Figure~\ref{fig:modelsize_scatter_appendix}, following the experiment described in Section~\ref{sec:quantitative}.
\newpage

\begin{figure}[!ht]
\centering
\subfloat[\label{fig:corr_bottle_1}\textit{traffic\_bottle} - Misweighting] {\includegraphics[scale=0.62]{figures/traffic/bottle_max_proxy_single.eps}}%
\hfill
\subfloat[\label{fig:corr_bottle_2}Correlation for Figure~\ref{fig:corr_bottle_1}] {\includegraphics[scale=0.62]{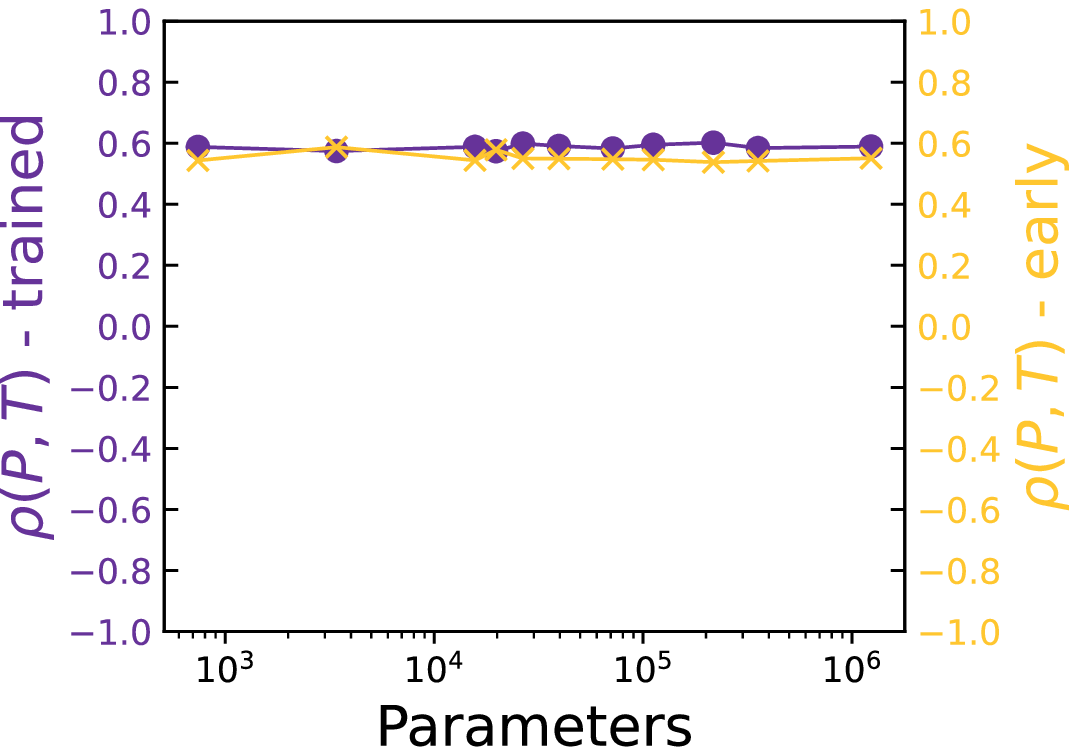}}%

\subfloat[\label{fig:corr_accel_1}\textit{traffic\_merge} - Misweighting] {\includegraphics[scale=0.62]{figures/traffic/accel_max_proxy_single.eps}}%
\hfill
\subfloat[\label{fig:corr_accel_2}Correlation for Figure~\ref{fig:corr_accel_1}] {\includegraphics[scale=0.62]{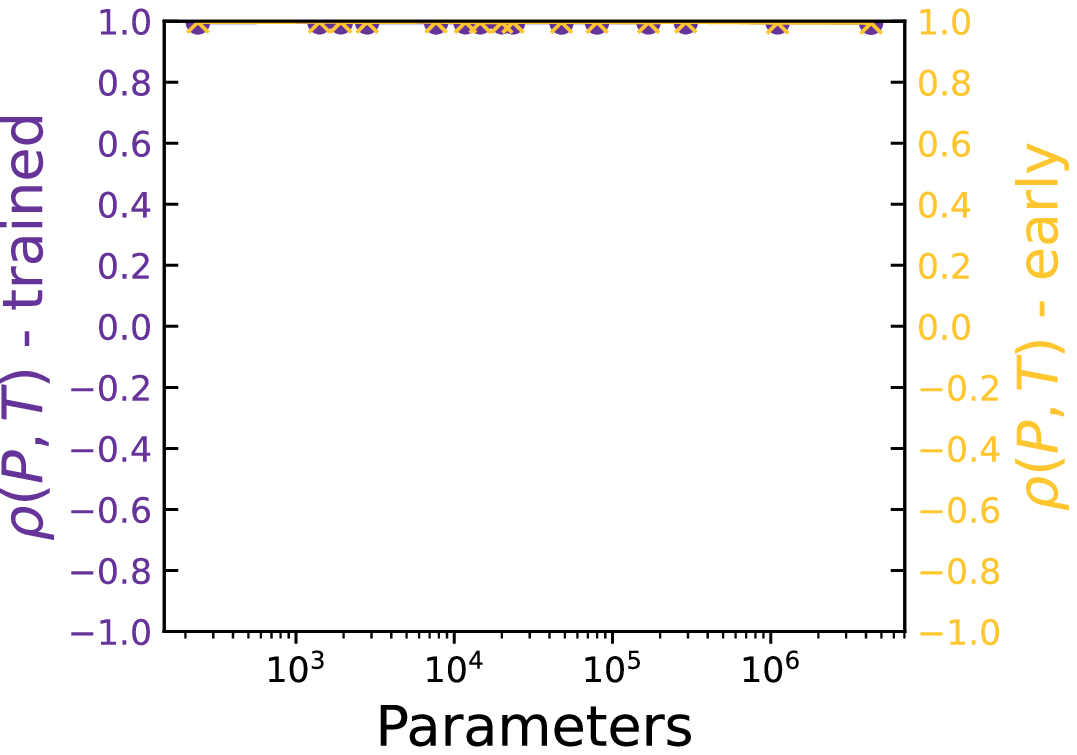}}%

\subfloat[\label{fig:corr_bus_1}\textit{traffic\_merge} - Ontological] {\includegraphics[scale=0.62]{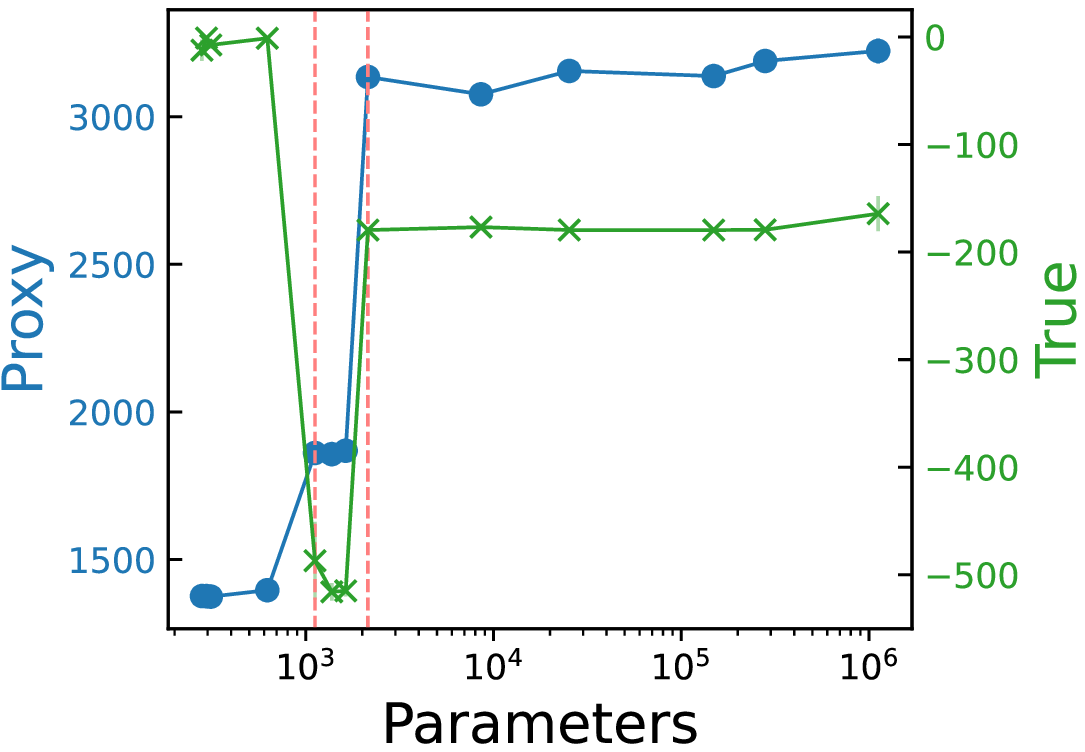}}%
\hfill
\subfloat[\label{fig:corr_bus_2}Correlation for Figure~\ref{fig:corr_bus_1}] {\includegraphics[scale=0.62]{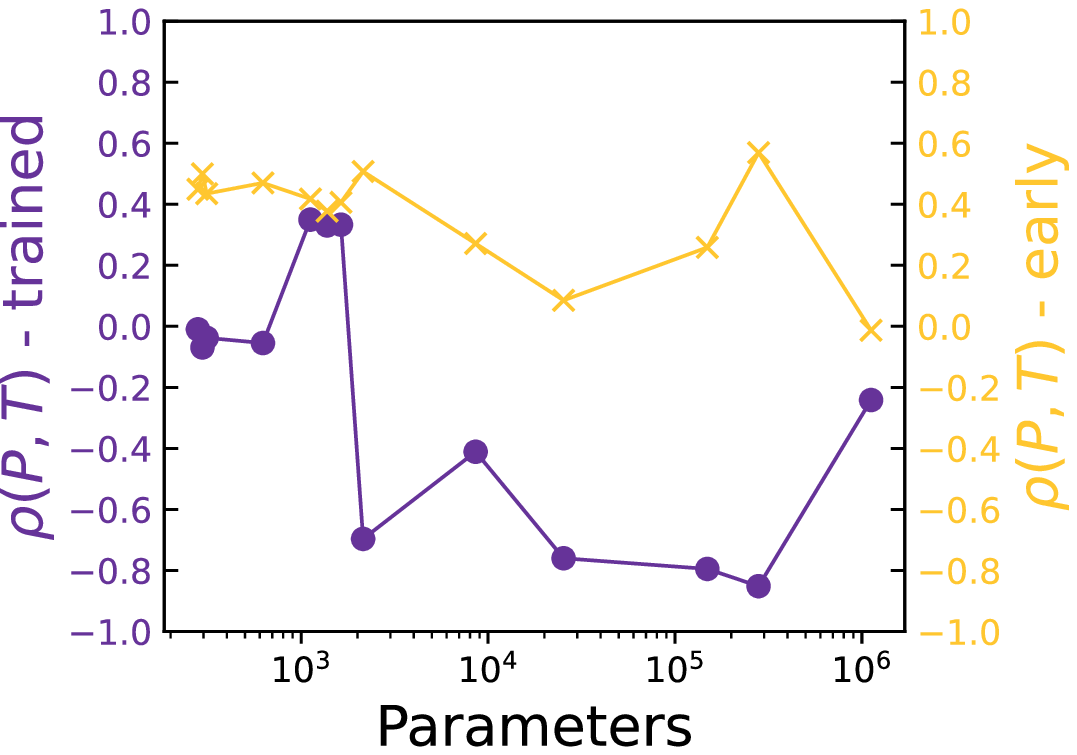}}%

\caption{Correlations between the proxy and true rewards, along with the reward hacking induced. In the left column, we plot the \textcolor{bluetext}{proxy reward} with ``$\bullet$" and the \textcolor{greentext}{true reward} with ``$\times$". In the right column, we plot the \textcolor{purpletext}{trained checkpoint correlation} and the \textcolor{yellowtext}{randomly initialized checkpoint correlation}.}
\label{fig:correlation_appendix}
\end{figure}

\subsection{Correlation between Proxy and True Rewards}\label{app:measurement_correlation}
We plot the correlation between proxy and true rewards, following the experiment described in Section~\ref{sec:correlation}. Interestingly, we see that reward hacking still occurs when there is positive correlation between the true and proxy rewards, e.g., in Figures~\ref{fig:corr_bottle_1}/\ref{fig:corr_bottle_2}. Unsurprisingly, proxy-true pairs which are highly correlated, e.g., Figure~\ref{fig:corr_accel_1}/\ref{fig:corr_accel_2} do not exhibit reward hacking. Finally, proxy-true pairs which are negatively correlated, e.g., Figure~\ref{fig:corr_bus_1}/\ref{fig:corr_bus_2} exhibit the most reward hacking.

\newpage
\newpage
\section{Polynomaly}

\subsection{Benchmark Statistics}
\label{app:benchmark_statistics}
See Table~\ref{tab:bench-stats} for Polynomaly's statistics. 
\begin{table*}[tbp]
\makebox[\textwidth][c]{
    \begin{tabular}{ccccc}
\toprule
Env. - Misspecification    &  \# Policies & \# Problematic & Rollout length  & Trusted policy size \\ \midrule
\trafficm~- misweighting  & $10$ & $7$ & $270$ &  $[96, 96]$\\
\trafficm~- scope & $16$ & $9$ & $270$ & $[16, 16]$ \\
\trafficm~- ontological  & $23$ & $7$ & $270$ & $[4]$ \\
\trafficb~- misweighting & $12$ & $9$ & $270$ & $[64, 64]$\\
COVID - ontological &  $13$& $6$& $200$ & $[16, 16]$ \\
\bottomrule
\end{tabular}%
}

\caption{Benchmark statistics. We average over 5 rollouts in traffic and 32 rollouts in COVID.}
\label{tab:bench-stats}%
\end{table*}%

\subsection{Receiver Operating Characteristic Curves} \label{app:roc}
We plot the ROC curves for the detectors described in Section~\ref{subsec:baseline_detector}. Our detectors are calculated as follows.

Let $P$ and $Q$ represent two probability distributions with $M = \frac{1}{2}(P + Q)$. Then the Jensen-Shannon divergence and the Hellinger distance between them is given by \begin{equation}\begin{aligned}\label{eqn:distance_metrics} \text{JSD}(P\vert\vert Q) \coloneqq \frac{1}{2} \text{KL}(P\vert\vert M) + \frac{1}{2}\text{KL}(Q \vert\vert M) \\ \text{Hellinger}(P, Q) \coloneqq \frac{1}{2} \int \left( \sqrt{dP} - \sqrt{dQ}\right)^2.
\end{aligned}\end{equation}

Our proposed detectors estimate the distance $\mathcal{D}(\pi_\text{trusted}, \pi_\text{unknown})$ between the trusted policy $\pi_\text{trusted}$ and unknown policy $\pi_\text{unknown}$ as follows: We generate $r$ rollouts of $\pi_\text{unknown}$, where $r = 5$ in the traffic environment and $r = 32$ in the COVID environment. Every $s$ steps of a rollout, where $s=10$ in the traffic environment and $s=1$ in the COVID environment, we set $P$ to be the action distribution of $\pi_\text{unknown}$ given the unknown agent's state at that timestep in the rollout and $Q$ to be the action distribution of $\pi_\text{trusted}$ given the unknown agent's state at that timestep in the rollout. Intuitively, if $P$ and $Q$ are far apart, then the trusted agent would have performed a different action than the unknown agent at that given timestep, indicating a possible case of reward hacking. We then compute either $\text{JSD}(P\|Q)$ or $\text{Hellinger}(P, Q)$ following Equation~\eqref{eqn:distance_metrics}. These distances are collected every $s$ steps over the entire rollout, and we calculate metrics on these distances (range, mean, etc.) to assign an anomaly score to the untrusted policy. 

\begin{figure}[htbp]
\centering
\subfloat[] {\includegraphics[scale=0.56]{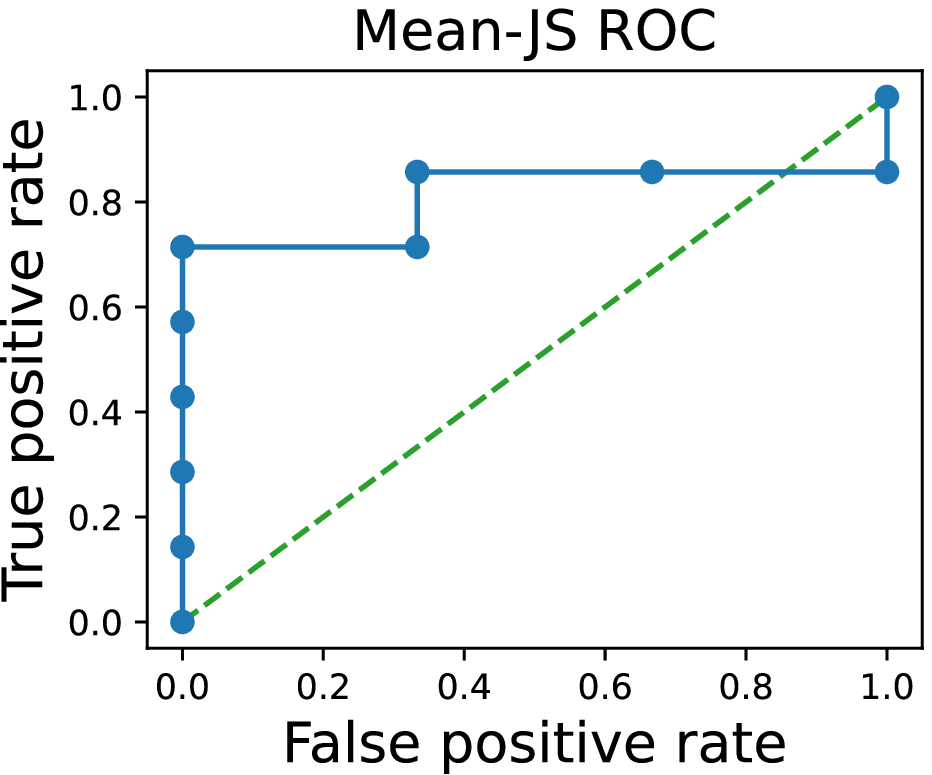}}%
\hfill
\subfloat[] {\includegraphics[scale=0.56]{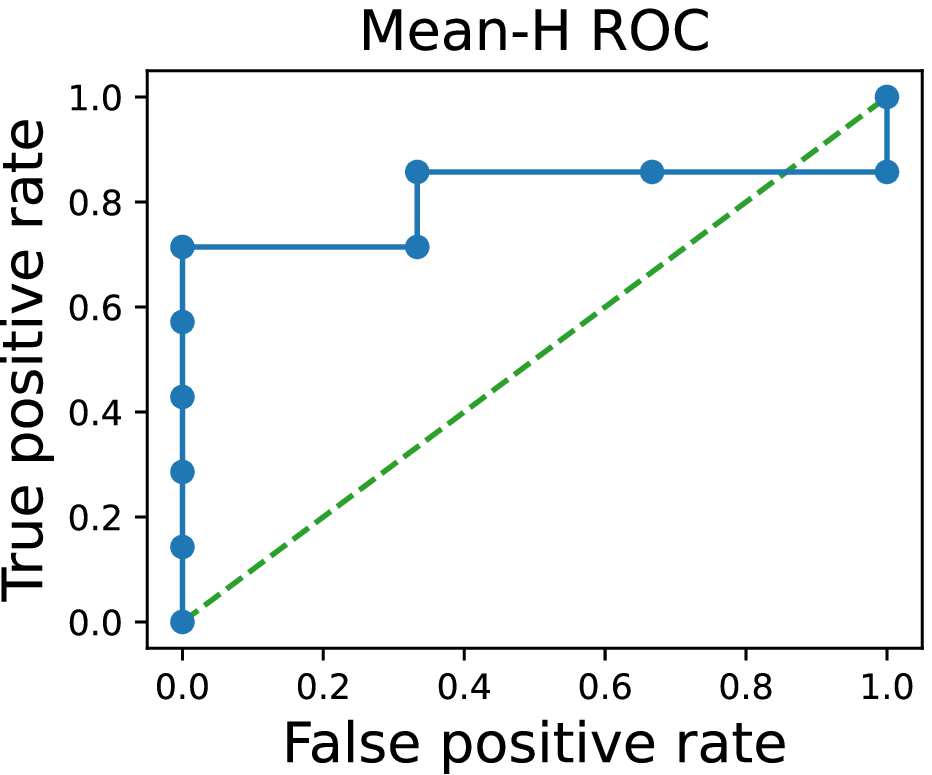}}%
\hfill
\subfloat[] {\includegraphics[scale=0.56]{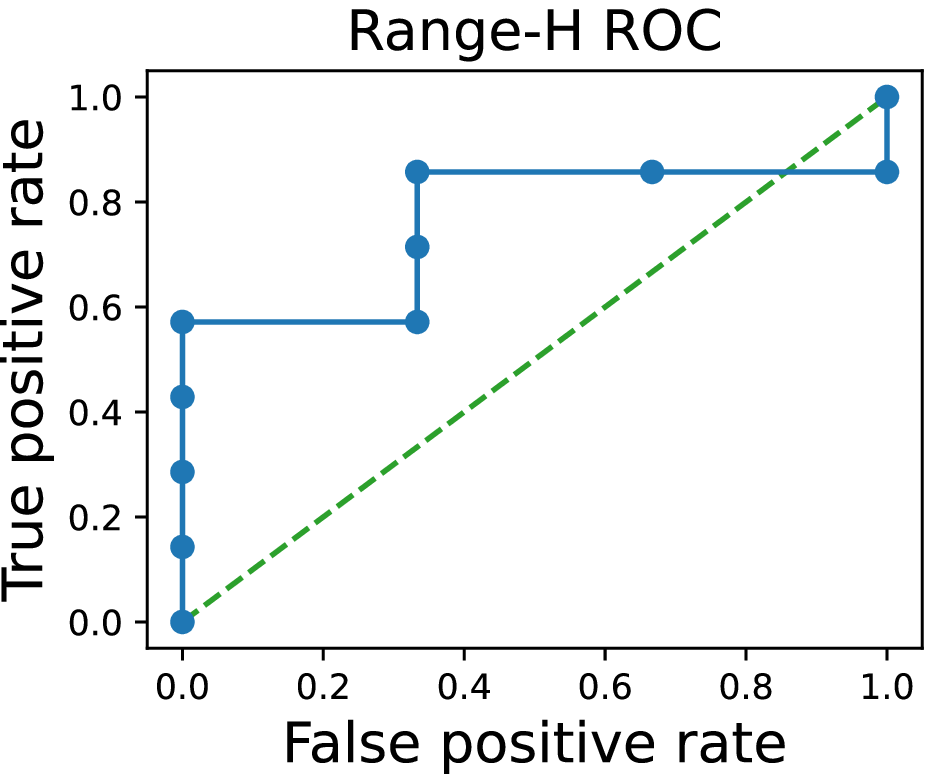}}%
\hfill
\caption{ROC curves for \trafficm~- misweighting.}
\end{figure}

\begin{figure}[htbp]
\centering
\subfloat[] {\includegraphics[scale=0.56]{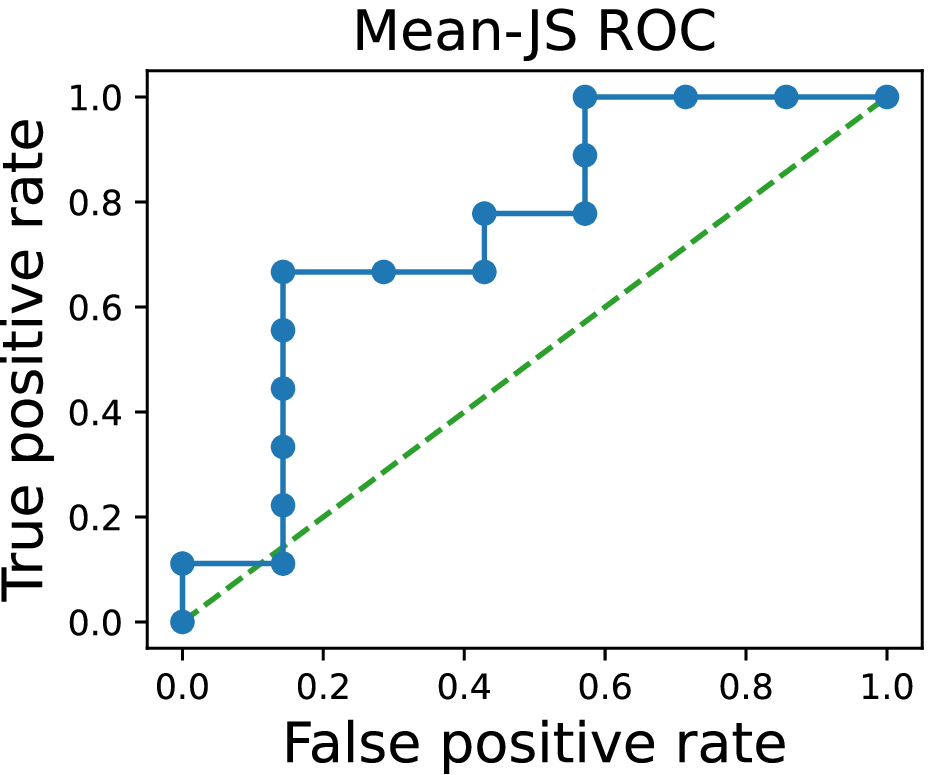}}%
\hfill
\subfloat[] {\includegraphics[scale=0.56]{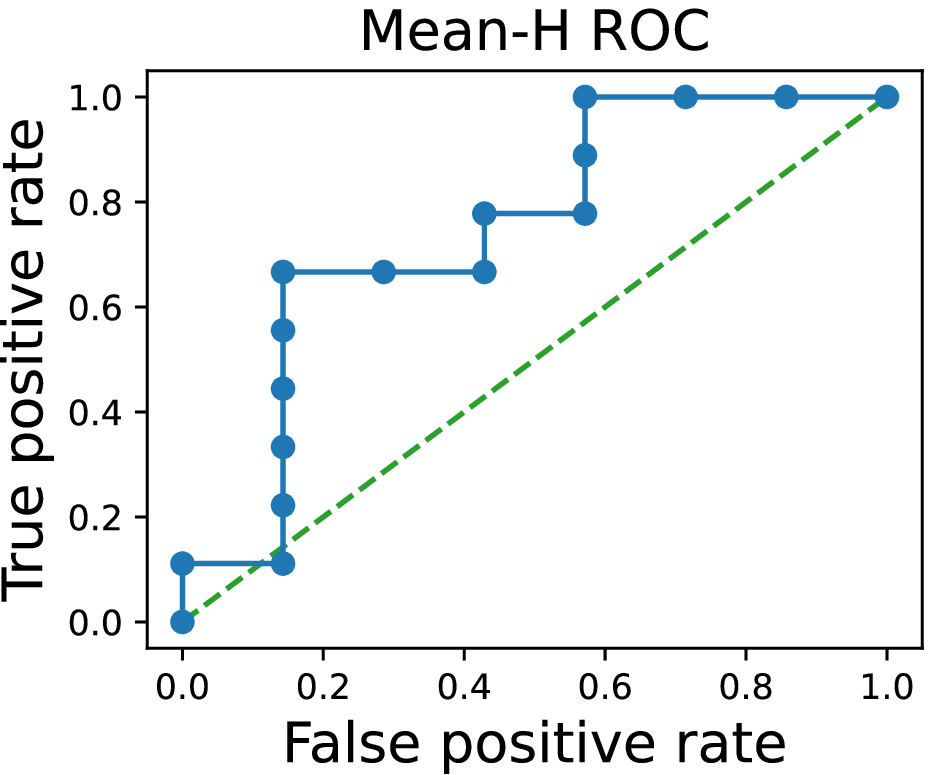}}%
\hfill
\subfloat[] {\includegraphics[scale=0.56]{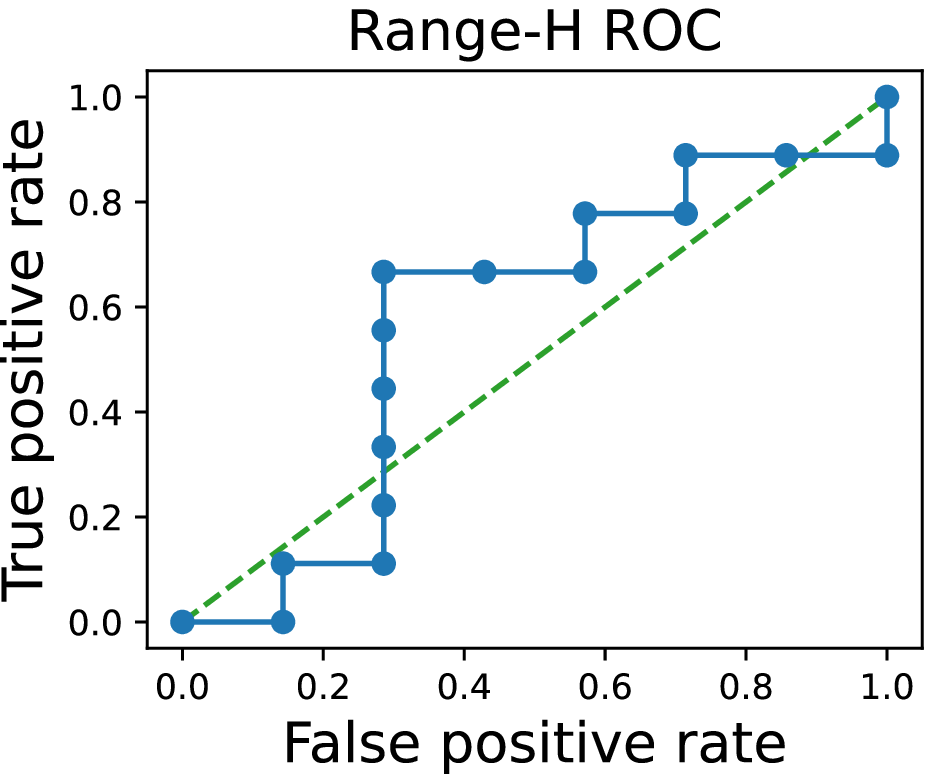}}%
\hfill
\caption{ROC curves for \trafficm~- scope.}
\end{figure}

\begin{figure}[htbp]
\centering
\subfloat[] {\includegraphics[scale=0.56]{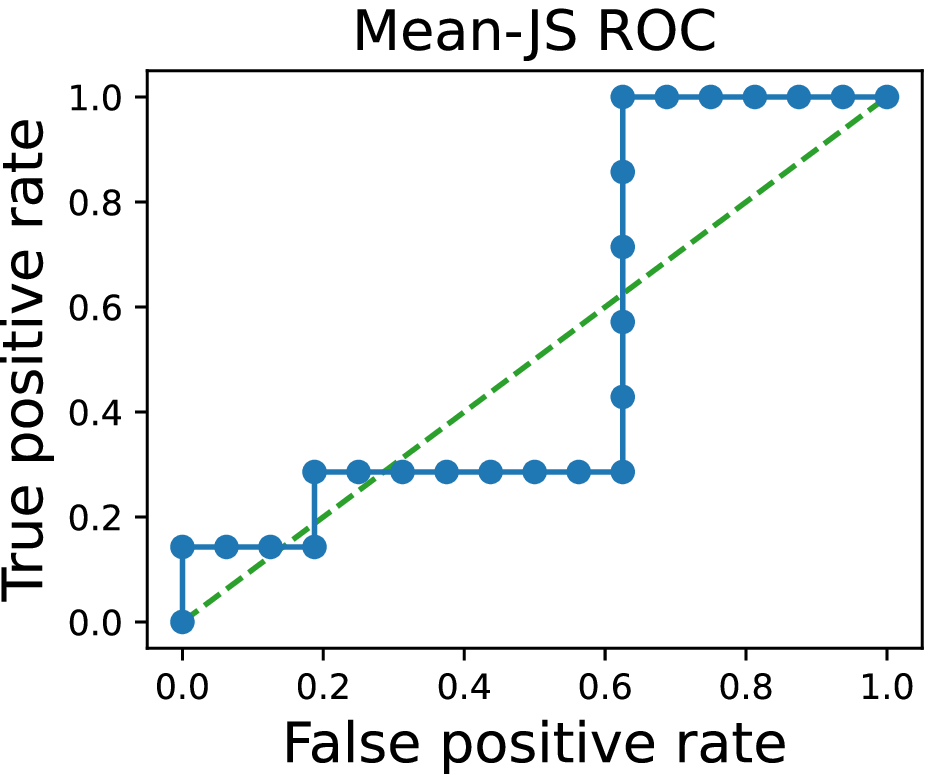}}%
\hfill
\subfloat[] {\includegraphics[scale=0.56]{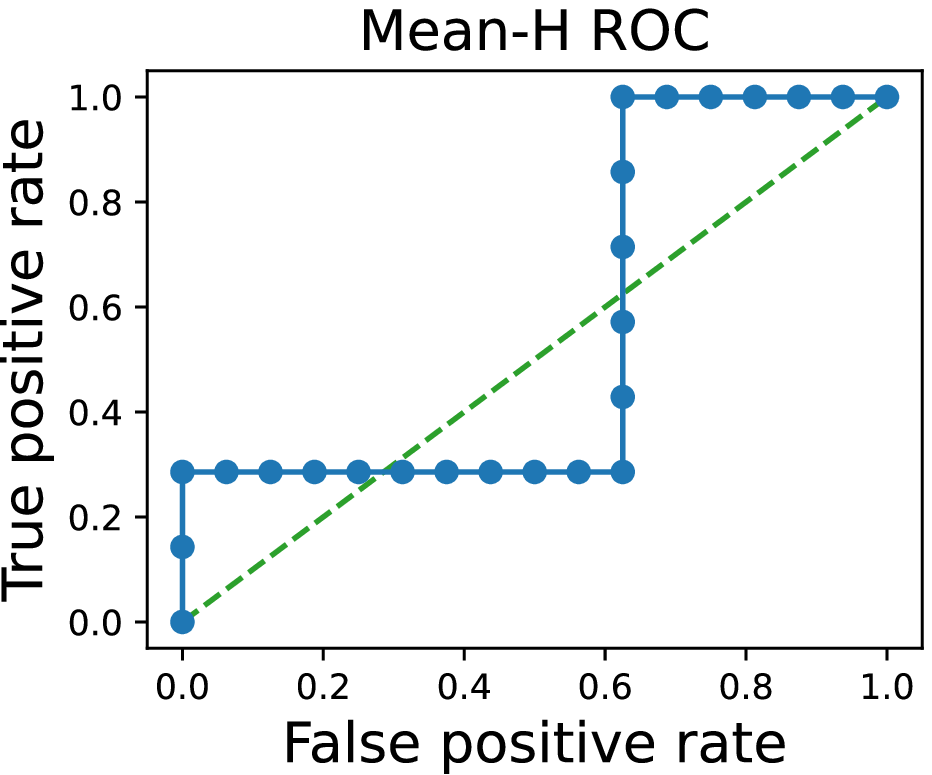}}%
\hfill
\subfloat[] {\includegraphics[scale=0.56]{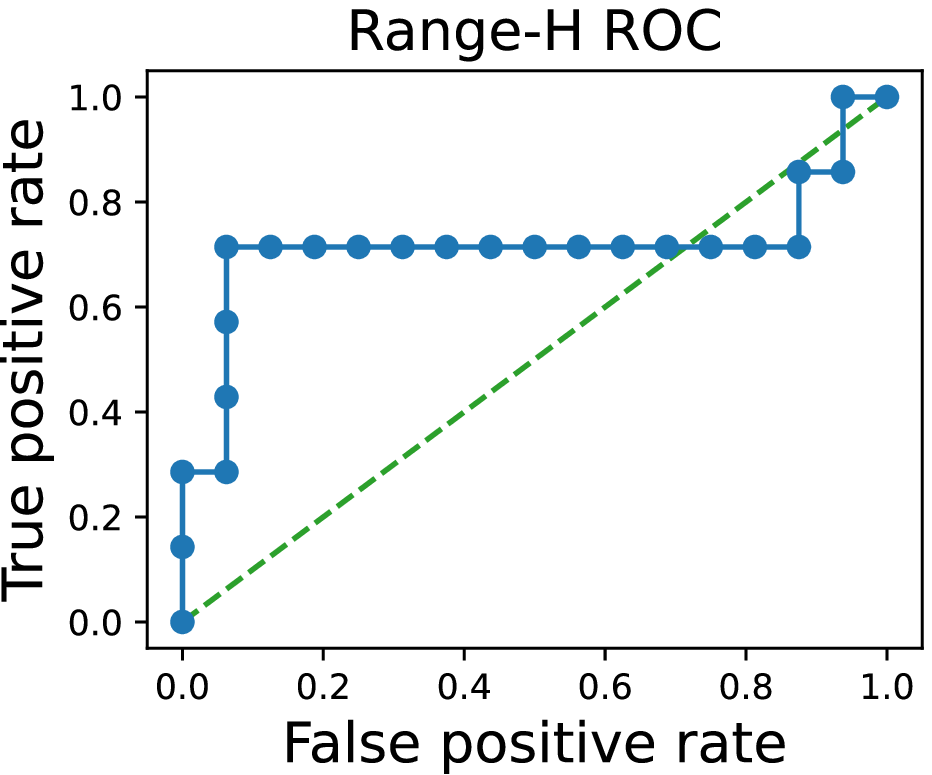}}%
\hfill
\caption{ROC curves for \trafficm~- ontological.}
\end{figure}

\begin{figure}[htbp]
\centering
\subfloat[] {\includegraphics[scale=0.56]{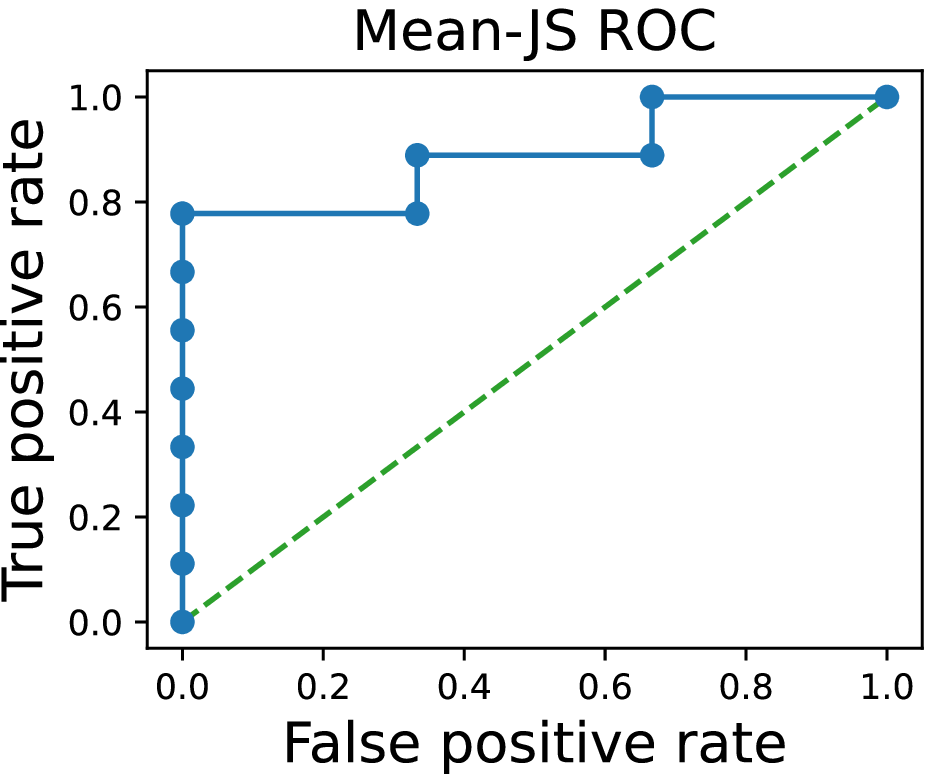}}%
\hfill
\subfloat[] {\includegraphics[scale=0.56]{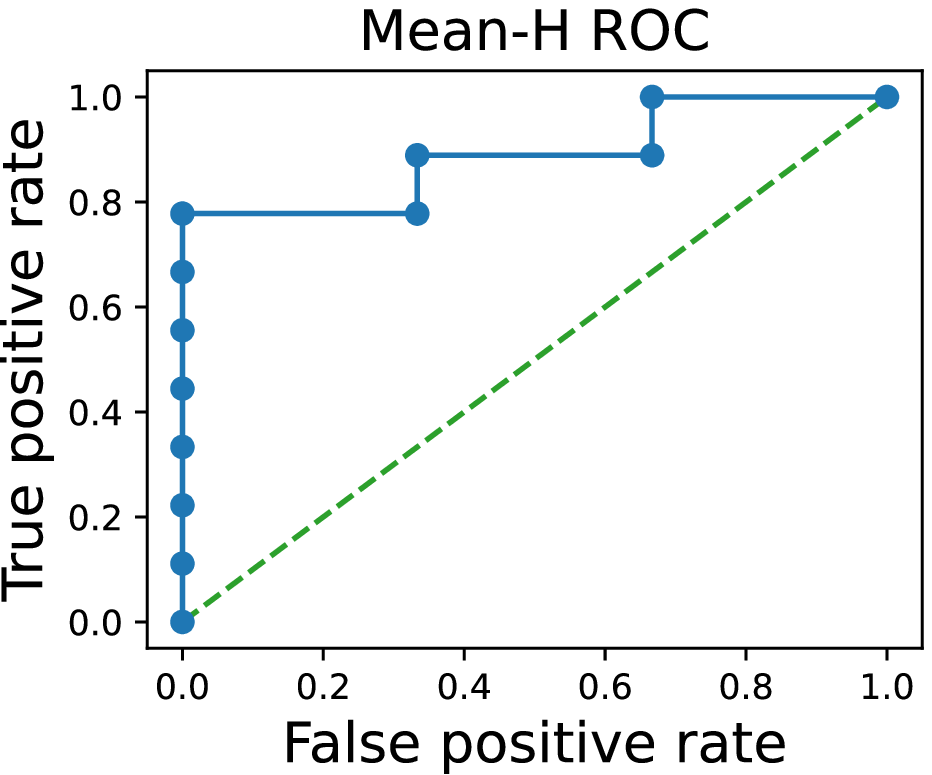}}%
\hfill
\subfloat[] {\includegraphics[scale=0.56]{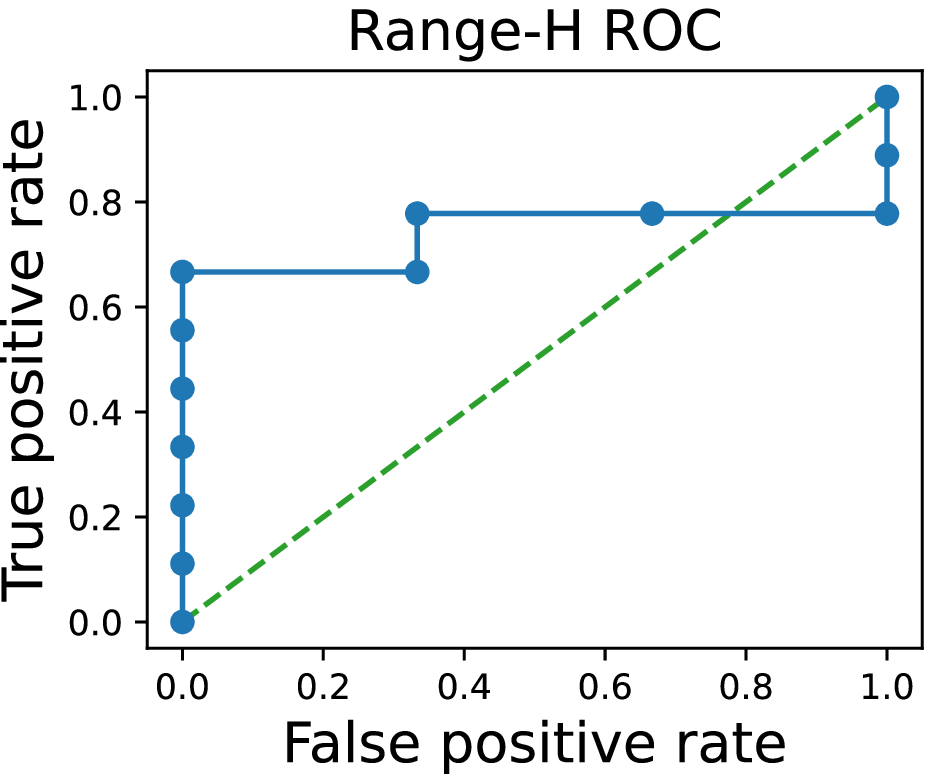}}%
\hfill
\caption{ROC curves for \trafficb~- misweighting.}
\end{figure}

\begin{figure}[htbp]
\centering
\subfloat[] {\includegraphics[scale=0.56]{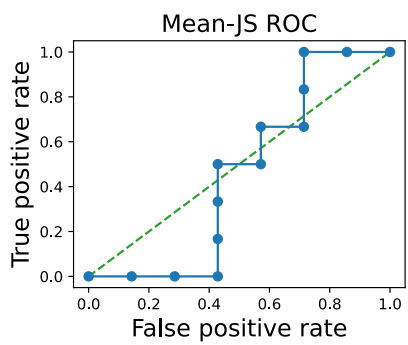}}%
\hfill
\subfloat[] {\includegraphics[scale=0.56]{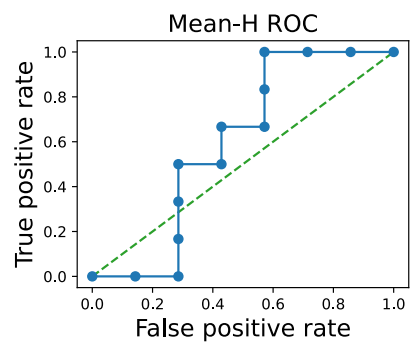}}%
\hfill
\subfloat[] {\includegraphics[scale=0.56]{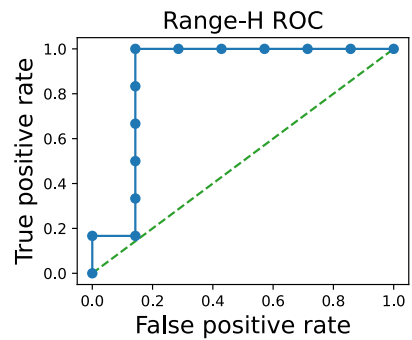}}%
\hfill
\caption{ROC curves for COVID - ontological.}
\end{figure}

\end{document}